\documentclass[twoside,11pt]{article}

\usepackage{jmlr2e}
\usepackage{color}

\usepackage{amssymb,amsmath}
\usepackage{bbm}

\usepackage{amsthm}
\usepackage{natbib}
\setcitestyle{sort,comma,authoryear}

\usepackage{helvet}
\usepackage[font=footnotesize,labelfont=bf]{caption}

\usepackage{xcolor}
\definecolor{myAmber}{rgb}{1, 0.749, 0.0}

\usepackage[ruled,linesnumbered]{algorithm2e}

\newtheorem{prop}{Proposition}

\newcommand{\1}{{\rm 1}\kern-0.24em{\rm I}}
\DeclareMathOperator*{\argmin}{arg\,min}
\newcommand{\bbP}{{\rm I}\kern-0.18em{\rm P}}
\newcommand{\bbR}{\mathbb{R}}

\ShortHeadings{Bridging asymmetric binary classification paradigms}{Li, Tong and Li}
\firstpageno{1}

\begin{document}

\title{Bridging Cost-sensitive and Neyman-Pearson Paradigms for Asymmetric Binary Classification}

\author{\name Wei Vivian Li \email vivian.li@rutgers.edu \\
       \addr Department of Biostatistics and Epidemiology\\
       School of Public Health \\
       Rutgers, The State University of New Jersey\\
       Piscataway, NJ 08854, USA
       \AND
       \name Xin Tong \email xint@marshall.usc.edu \\
       \addr Department of Data Sciences and Operations\\
       Marshall School of Business\\
       University of South California\\
       Los Angeles, CA 90089, USA
       \AND
       \name Jingyi Jessica Li \email jli@stat.ucla.edu \\
       \addr Department of Statistics \\
       University of California, Los Angeles\\
       Los Angeles, CA 90095, USA
       }

\editor{}

\maketitle

\begin{abstract}
Asymmetric binary classification problems, in which the type I and II errors have unequal severity, are ubiquitous in real-world applications. To handle such asymmetry, researchers have developed the cost-sensitive and Neyman-Pearson paradigms for training classifiers to control the more severe type of classification error, say the type I error. The cost-sensitive paradigm is widely used and has straightforward implementations that do not require sample splitting; however, it demands an explicit specification of the costs of the type I and II errors, and an open question is what specification can guarantee a high-probability control on the population type I error. In contrast, the Neyman-Pearson paradigm can train classifiers to achieve a high-probability control of the population type I error, but it relies on sample splitting that reduces the effective training sample size. Since the two paradigms have complementary strengths, it is reasonable to combine their strengths for classifier construction. In this work, we for the first time study the methodological connections between the two paradigms, and we develop the TUBE-CS algorithm to bridge the two paradigms  from the perspective of controlling the population type I error. 
\end{abstract}


\section{Introduction}

Asymmetric binary classification problems, where the consequence of misclassifying observations in one class is more severe than the other class, are ubiquitous in real-world applications.
For example, in medical diagnosis, misclassifying  a malignant tumor biopsy sample as benign is more severe than misclassifying a benign sample as malignant \citep{mazurowski2008training,park2011cost,moon2012computer}; in email spam detection, removing a non-spam email leads to more severe consequences than missing a spam email \citep{carreras2001boosting,zhou2014cost}; in political conflict prediction, missing a conflict has more critical consequences than vice versa \citep{beck2000improving,cederman2017predicting}.
In addition to these examples, asymmetric binary classification problems exist in geologic studies \citep{horrocks2015classification,fernandez2017large}, fraud detection \citep{sahin2013cost,bahnsen2013cost}, and other medical diagnosis and prognosis problems \citep{artan2010prostate,ali2016can}.

Throughout this article, we refer to the two classes as class 0 and class 1, and we use class 0 to represent the more important class.
Accordingly, in the above examples, malignant biopsy samples, non-spam emails, and political conflict events are class 0 observations.
With this encoding in asymmetric classification, the type I error, defined as the conditional probability of misclassifying a class 0 observation into class 1, demands a higher priority than the type II error, the conditional probability of misclassifying a class 1 observation into class 0.
However, the asymmetric importance of type I and II errors are not reflected in common classification practices that aim to minimize the overall classification error or maximize the area under the receiver-operating characteristic curve. As a result, asymmetric classification paradigms have been developed.

To date, asymmetric classification has two major paradigms: cost-sensitive (CS) and Neyman-Pearson (NP). The CS paradigm incorporates two misclassification costs, with the type I error cost greater than the type II error cost, into its objective function so that the classifier minimizing the objective function would have the type I error smaller than the type II error. In contrast, the NP paradigm does not use misclassification costs but instead minimizes a constrained objective function, where the type II error is to be minimized subject to an upper bound on the type I error.

Desirably, the objective functions of the two paradigms have one-to-one correspondence at the population level: when both paradigms' objective functions are defined by the population type I and II errors, the two functions' minimizers are the same oracle classifier, whose population type I error is equal to the type I error upper bound in the NP paradigm. However, such correspondence is not always achievable at the algorithm level because the two paradigms may use different algorithms to construct practical classifiers from training data, and practical classifiers may not mimic the oracle classifier to the same extent. Hence, in practice, the connection between the two paradigms is unclear, so is the paradigm choice for a particular application. Another question is whether the two paradigms' comparative advantages can be combined into a practical classifier's construction. 

This article addresses these questions and is structured as follows. In Sections \ref{sec: CS} and \ref{sec:NP}, we review the CS and NP classification paradigms, respectively; in detail, for each paradigm, we explain its statistical framework, its implementation of the type I error priority, and its comparative advantages. In Section \ref{sec:exact-compare}, we discuss two special implementations of the CS paradigm that are equivalent to the NP paradigm in constructing practical classifiers. In Section \ref{sec:gen-alg}, we develop the TUBE algorithm for estimating the type I error upper bound---the key property of the NP paradigm---of a practical classifier constructed under the CS paradigm; the algorithm establishes a link between the two paradigms in practice. In Section \ref{sec:simulation}, we propose the TUBE-CS algorithm for selecting the type I error cost parameter in the CS paradigm given a target type I error upper bound. Then we use simulation to verify the performance of the TUBE-CS algorithm. In Section \ref{sec:realdata}, we use real data studies to verify that the TUBE-CS algorithm outperforms a vanilla CS algorithm commonly used in practice. The Appendix includes proofs, a supplementary algorithm, and supplementary figures.


\section{The CS classification paradigm}\label{sec: CS}

First, we introduce notations used throughout of the paper.  Let $(X, Y)$ be a random pair in which $X \in \mathcal{X}\subset \bbR^d$ represents $d$ features and $Y\in\{0,1\}$ encodes a binary class label associated with $X$. A classifier $\phi(\cdot)$ is a map $\phi:\mathcal{X}\rightarrow\{0,1\}$ that outputs a predicted class label given the input features $X$. The population type I error of $\phi(\cdot)$ is $R_0(\phi) = \bbP(\phi(X)=1|Y=0)$, and the population type II error is $R_1(\phi) = \bbP(\phi(X)=0|Y=1)$. With the priors, i.e., proportions of classes $0$ and $1$ in the population, which are denoted respectively by $\pi_0=\bbP(Y=0)$ and $\pi_1=\bbP(Y=1)$ (such that $\pi_0+\pi_1=1$), the overall classification error of $\phi(\cdot)$ is
\begin{eqnarray}\label{eqn:classical objective}
R(\phi) = \bbP(\phi(X) \neq Y) =  \pi_0R_0(\phi) + \pi_1R_1(\phi)\;.
\end{eqnarray}
We define the classical paradigm as the one that seeks a classifier to minimize $R(\cdot)$. The classical paradigm prioritizes the type I error based on $\pi_0$ (larger $\pi_0$ gives the type I error a higher priority) and does not allow a user-specified upper bound on the type I error. Hence, it is unsuitable for asymmetric classification.

The CS paradigm is a natural modification of the classical paradigm for asymmetric classification. It assigns explicit misclassification costs $c_0$ and $c_1$ to the type I and II errors respectively (Table \ref{tab:cost-mat}) \citep{margineantu2000does, margineantu2002class, sun2007cost}.
Then it seeks a classifier to minimize\footnote{In some research papers, the objective function of the CS paradigm is 
$R_\text{CS}(\phi) = c_0\pi_0R_0(\phi) + c_1\pi_1R_1(\phi)$.}
\begin{eqnarray}\label{eqn:cs objective}
R_\text{CS}(\phi) = c_0R_0(\phi) + c_1R_1(\phi)\,.
\end{eqnarray}
In certain applications, $c_0$ and $c_1$ can be specified in an objective manner, such as monetary and time costs \citep{turney2002types}. However, in most applications, users cannot concretize $c_0$ and $c_1$ as real-world costs. For example, it is ethically impossible to specify the cost of misdiagnosing a cancer patient as undiseased \citep{vidrighin2008proicet}. Due to this challenge, users in practice specify $c_0$ and $c_1$ in a data-driven way: they vary $(c_0, c_1)$, train a classifier for each pair of costs, evaluate the classifier's empirical type I and II errors on evaluation data, and finally choose the pair of costs so that the empirical type I error is below and closest to their target type I error upper bound $\alpha$. While this approach is intuitive and easy to implement, it has a critical drawback---it does not inform whether the classifier trained with the chosen $(c_0, c_1)$ has the population type I error under $\alpha$, a question the NP paradigm can answer (Section~\ref{sec:NP}). Before introducing the NP paradigm, we summarize the CS paradigm's practical implementations and their relation to the type I error control.

%

\begin{table}[tbh!]
\caption{\label{tab:cost-mat}The misclassification cost matrix.}
\centering
  \begin{tabular}{| l | c | c |}
    \hline
     & true class is 0 & true class is 1 \\ \hline
    predicted class is 0 & 0 & $c_1$ \\ \hline
    predicted class is 1 & $c_0$ & 0 \\
    \hline
  \end{tabular}
\end{table} 

\subsection{A review of practical implementations of the CS paradigm}

We summarize practical implementations of the CS paradigm into three categories (Table \ref{tab:cs-summary}): (a) pre-training approaches, which modify training data to reflect misclassification costs; (b) in-training approaches, which change the internal algorithm of a classification method (e.g., logistic regression) to incorporate misclassification costs; (c) the post-training approach, which incorporates misclassification costs into the threshold on the posterior probability predicted by a trained classification method. 

\begin{table}[tbh!]
\caption{\label{tab:cs-summary}Three categories of practical implementations of the CS classification paradigm.}
\includegraphics[width = \textwidth]{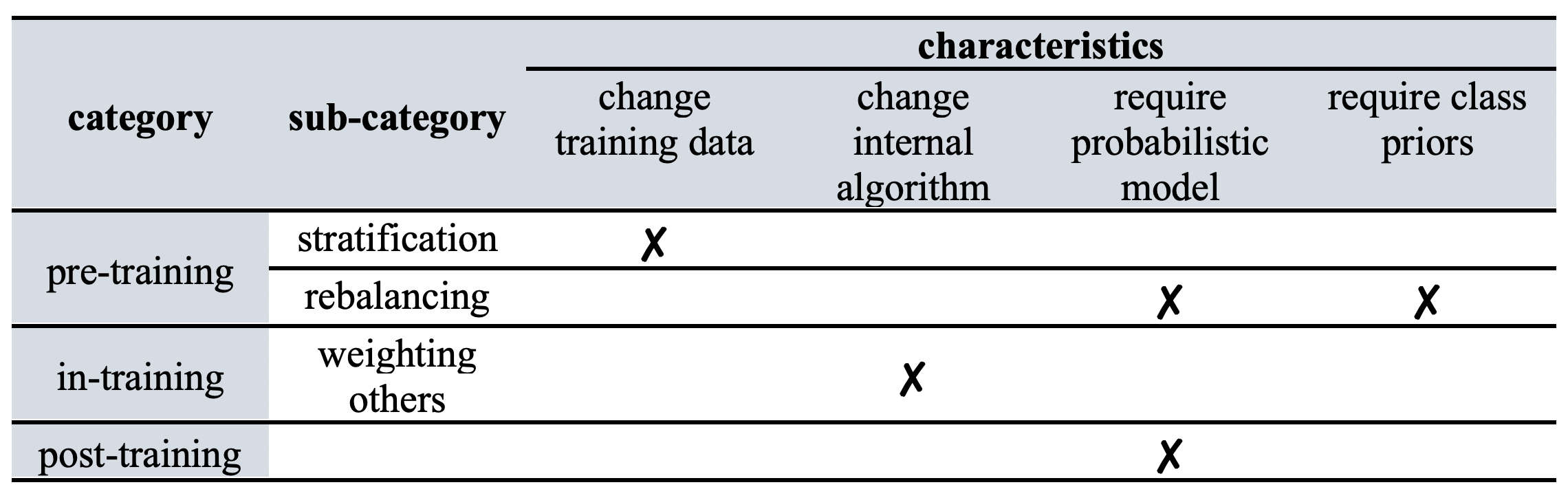}
\end{table} 

\textbf{(a) Pre-training approaches}. They consist of two sub-categories: stratification and rebalancing approaches. 
Stratification approaches change the proportions of the two classes in the training data such that the more important class $0$ has a larger proportion in the training data than in the population  \citep{zadrozny2003cost}. The implementation is to either downsample the class $1$ or oversample the class $0$ \citep{margineantu2002class,webb2005application,pelayo2007applying,pelayo2012evaluating}. Stratification approaches have been applied to credit card fraud detection \citep{chan1998toward} and studied for their effects on decision-tree-based classification methods \citep{drummond2000exploiting}.
In contrast, the rebalancing approach does not change the training data but replaces class priors $\pi_0$ and $\pi_1$ respectively by normalized $c_0$ and $c_1$ (such that the two normalized costs add up to $1$); however, it is only applicable to probabilistic classification methods, in which class priors are no longer estimated from training data but set to the specified $c_0$ and $c_1$. 
For example, the rebalancing approach has been implemented with the linear discriminant analysis method (function \verb|lda| in the R package \verb|MASS| \citep{mass}) and the na\"ive Bayes method (function \verb|naive_bayes| in the R package \verb|naivebayes| \citep{naivebayes}).  

\textbf{(b) In-training approaches}. They change the internal algorithm of a classification method. 
A popular in-training approach is weighting, which assigns individual training observations with weights proportional to their misclassification costs ($c_0$ and $c_1$ for class $0$ and $1$ observations, respectively) during the training of a classification method. 
This weighting approach has been implemented with classification methods including the logistic regression (function \verb|glm| in the R package \verb|stats| \citep{stats}), the penalized logistic regression (function \verb|glmnet| in the R package \verb|glmnet| \citep{glmnet}), and the support vector machine (function \verb|svm| in the R package \verb|e1071| \citep{e1071}). A technical note is that, in the above three R functions, assigning an integer weight $w$ to the class $0$ observations and a weight of $1$ to the class $1$ observations is equivalent to replicating each class $0$ observation $w$ times before regular training.
Unlike the weighting approach, most other in-training approaches are specific to a classification method and not generalizable. 
For instance, tree-based classification methods, adaptive boosting (AdaBoost), and neural networks  all have special in-training approaches. Among tree-based methods, the classification and regression tree (CART) method incorporates misclassification costs into the splitting criterion used in tree growing \citep{breiman2017classification}; the random forest method also implements this approach in the weighted random forest (available in the R package \verb|randomForest|) \citep{chen2004using}; Bradford et al. discussed different pruning algorithms for decision trees to reflect asymmetric misclassification costs \citep{bradford1998pruning}. For AdaBoost, a variant Adacost was proposed to incorporate asymmetric misclassification costs into the weighting of data points \citep{fan1999adacost}; \cite{sun2007cost} discussed three modifications of the weight-update formula in AdaBoost to incorporate misclassification costs.
For neural networks, \cite{kukar1998cost} modified the back-propagation learning algorithm to incorporate misclassification costs; \cite{zhou2006training} proposed ensemble learning to combine multiple cost-sensitive neural network classifiers.

\textbf{(c) Post-training approach}. It requires a classification method to assign each test observation an estimated posterior probability of being in class $1$.   
What it does is to select a threshold on the estimated probability so that each test observation is classified as $0$ or $1$ \citep{duda1973pattern}. The threshold selection depends on misclassification costs. This post-training approach has a notable difference from the previous pre-training and in-training approaches: it does not alter either the training data or the internal algorithm of a classification method. In other words, it applies to the output of a classification method without requiring access to the training data or the implementation of the classification method. An example is the MetaCost method, which constructs an ensemble classifier that implements this post-training approach \citep{domingos1999metacost}.   

\subsection{Specification of misclassification costs}\label{sec: cs-spec}
In the aforementioned three categories of CS implementation approaches, users must specify the misclassification costs. 
However, there lacks consensus on how to specify the costs objectively.
In some applications, the costs can be specified by domain experts, such as in the prediction of head injury recovery \citep{elder1996machine,liu2016applying}, the diagnosis of heart disease \citep{king1995statlog}, and the classification of customer credit \citep{elder1996machine}. However, in most applications, costs cannot be specified by domain experts but need to be set either arbitrarily or by some prediction criteria. For instance, many studies in medical diagnosis \citep{schaefer2007cost,vidrighin2008proicet,park2011cost,alizadehsani2012diagnosis}, food safety classification \citep{liu2016applying}, and climate change prediction \citep{lu2008ground} have used one or multiple set(s) of arbitrarily specified costs. 
In addition to being arbitrarily chosen, the costs may be chosen to optimize a classification accuracy measure on evaluation data.
For example, \cite{sun2007cost} selected the costs that led to the largest $F$ score (the harmonic mean of precision and recall) in several medical diagnose problems; \cite{lan2010investigation} compared the costs in terms of the overall and class-specific misclassification errors in thyroid level prediction; \cite{krawczyk2015hybrid} selected the costs to optimize the receiver operating characteristic (ROC) curves in breast thermogram classification.

\subsection{Population type I error in CS classification}\label{sec: cs-sim}
The CS classification paradigm is popular in practice due to its easy implementation. 
However, it remains unclear whether and to what extent this paradigm can control the population type I error, i.e., the more severe type of classification error. 
Here, we study a CS classifier's performance from the perspective of controlling the population type I error.
Without loss of generality, we assume $c_0+c_1=1\ (c_0, c_1>0)$  in the following text.

We denote a practical CS classifier by $\hat\phi_c$, in which $c$ is the  type I error cost (i.e., $c=c_0)$.
Suppose that one wishes to find $c$ such that $\hat\phi_c$ has the population type I error under a specified level $\alpha$. A natural but na\"ive (``vanilla") implementation of the CS paradigm is to solve 
\begin{eqnarray} 
\min\limits_{\hat\phi_c} \widehat R_1(\hat\phi_c) \text{ subject to } \widehat R_0(\hat\phi_c) \le \alpha\;,
\end{eqnarray} 
in which  $\widehat R_0(\hat\phi_c)$ and $\widehat R_1(\hat\phi_c)$  are the empirical type I and II errors of  $\hat\phi_c$ on evaluation data, respectively. 

We detail this vanilla implementation in Algorithm \ref{alm:cs}, which is adaptive to any CS implementation approach (pre-training, in-training, or post-training).
Specifically, the training data are randomly divided into two subsets: one 
to train CS classifiers corresponding to candidate costs $c_{0,1} < c_{0,2} < \dots < c_{0,I}$; 
the other to evaluate the empirical type I error of each classifier. 
Among the $I$ CS classifiers, the chosen one corresponds to the smallest cost such that the empirical type I error is no greater than $\alpha$. 

\begin{algorithm}[htb!]
\DontPrintSemicolon
\caption{The vanilla CS implementation to control the type I error
\label{alm:cs}}
\SetKw{KwBy}{by}
\SetKwInOut{Input}{Input}\SetKwInOut{Output}{Output}
\SetAlgoLined

\Input{$\mathcal{S}=\mathcal{S}^0\cup \mathcal{S}^1$: training data ($\mathcal{S}^0$ for class $0$ and $\mathcal{S}^1$ for class $1$)\\
$\alpha$: target upper bound on the type I error\\
$c_{0,1} < c_{0,2} < \dots < c_{0,I}$: candidate type I error costs\\
}
$\mathcal{S}^0_1, \mathcal{S}^0_2 \leftarrow$ randomly split $\mathcal{S}^0$ into two subsets  

\For{$i\gets1$ \KwTo $I$}{
$\hat\phi_i(\cdot)=$ CS-classifier($\mathcal{S}^0_1\cup\mathcal{S}^1$, $c_{0,i}$) 

$r_{0,i}=\widehat R_0(\hat\phi_i)$\tcp*{empirical type I error on $\mathcal{S}^0_2$}
}
$ i^* = \left\{ \begin{array}{ll}
 	\min\{i: r_{0,i}\leq \alpha\} & \text{ if } r_{0,I}\le\alpha\\
 	I & \text{ otherwise}
 \end{array}
 \right.$
 
\Output{$\hat{\phi}^{\textrm{vanilla-CS}}(\cdot)=\hat{\phi}_{i^*}(\cdot)$ \tcp*{the vanilla-CS classifier}}
\end{algorithm}


\begin{figure}[tbh!]
\centering
\includegraphics[width = .9\textwidth]{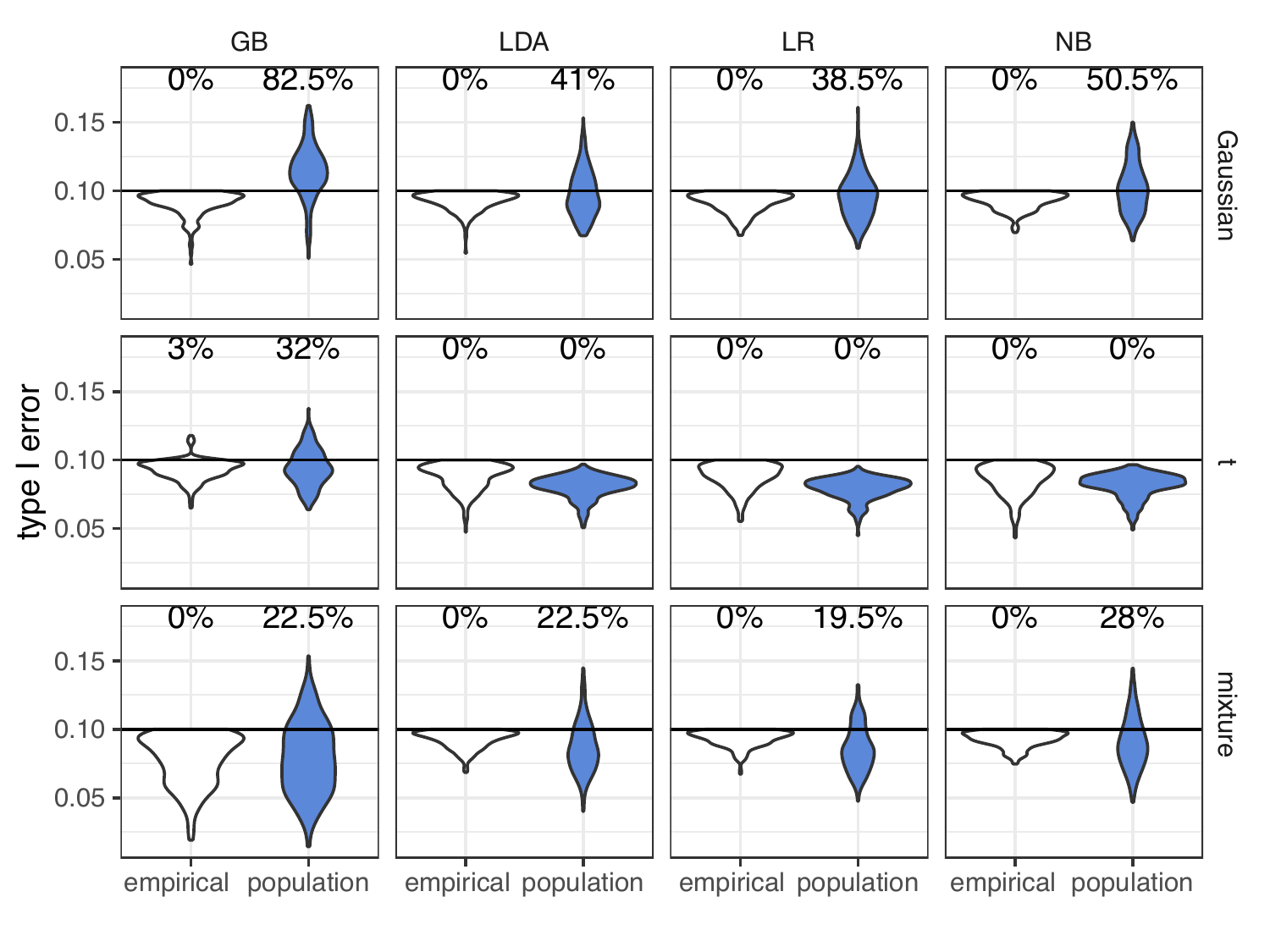}
\caption{\label{fig:s1-CS}
The empirical type I errors and population type I errors of CS classifiers trained by Algorithm \ref{alm:cs}. We set the target type I error upper bound $\alpha = 0.1$ and use the stratification approach combined with four classification methods: gradient boosting (GB), linear discriminant analysis (LDA), logistic regression (LR), and na\"ive Bayes (NB). We consider three data distributions: Gaussian, $t$, and Mixture. The violation rate of type I errors (the frequency of type I errors being greater than $\alpha = 0.1$) in each case is summarized as a percentage and marked on top of each violin plot.
}
\end{figure}

We use a simulation study to demonstrate that Algorithm \ref{alm:cs} cannot guarantee to control the population type I error of its chosen classifier under $\alpha$. In this simulation, we use the stratification approach, a type of pre-training approach, as an example to show that the population type I error of a CS classifier constructed by Algorithm \ref{alm:cs} is not bounded by $\alpha$ with high probability, even though the empirical type I error of the classifier is under $\alpha$.
We implement the stratification approach with four base classification methods: logistic regression, gradient boosting, linear discriminant analysis, and na\"ive Bayes. 
We consider three data distributions, denoted as Gaussian, Multivariate $t$, and Mixture.
\begin{enumerate}
  \item Gaussian. $\bbP(Y=0)=0.5$. When $Y = 0$, $X_{d\times 1} \sim N(\mu_0, \Sigma_0)$; when $Y = 1$, $X_{d\times 1} \sim N(\mu_1, \Sigma_1)$. The means are $\mu_0=(0,0,0,\dots,0)^{\top}$ and $\mu_1 = (1.5,1.5,0,\dots,0)^{\top}$. The covariance matrices are $\Sigma_0= I_{d}$ and $\Sigma_1$ with $1$ in  diagonal entires,  $0.5$ in superdiagonal and subdiagonal entries, and $0$ in the rest of entries.
  \item Multivariate $t$. $\bbP(Y=0)=0.5$. When $Y=1$, $X_{1:2}\sim t_3(\mu_0)$ and $X_{3:d}\sim N( 0,  I_{d-2})$; when $Y=0$, $X_{1:2}\sim t_3(\mu_1)$ and $X_{3:d}\sim N( 0,  I_{d-2})$. The noncentrality parameters are $\mu_0 = (0,0)^{\top}$ and $\mu_1 = (2.5, 2.5)^{\top}$.
  \item Mixture. $\bbP(Y=0)=0.5$. When $Y=0$, $X \sim \frac{1}{2}N(\mu_{1}, I_d)+\frac{1}{2}N(\mu_{2},  I_d)$; when $Y=1$, $X \sim N(\mu_1, I_d)$. The mean parameters are $\mu_1 = (a,a,\dots,a)^{\top}$ and $\mu_2 = (-a,-a,\dots,-a)^{\top}$, where $a = 2/\sqrt{d}$.
\end{enumerate}

For each distribution, we set $d=30$ and simulate a training dataset with size $n_1 = 1{,}000$ and a large evaluation dataset with size $n_2 = 1{,}000{,}000$ (to approximate the population). In each simulation, we use Algorithm \ref{alm:cs} (with candidate type I error costs ranging from $0.51$ to $0.99$ in a step size of $0.02$) to train a CS classifier that has the empirical type I error under $\alpha = 0.1$. Then we approximate the population type I error of the trained classifier using the large evaluation dataset. We repeat the simulation for $200$ times to calculate the violation rates of the empirical type I errors and the population type I errors, respectively. Note that the violation rate is defined as the frequency of type I errors exceeding $\alpha$.  Figure \ref{fig:s1-CS} summarizes the results and shows that the CS classifiers constructed by Algorithm \ref{alm:cs} fail to have population type I errors under $\alpha$ with high probability (i.e., the violation rates are large), even though these classifiers have empirical type I errors strictly\footnote{Note that there is a $3\%$ violation rate of empirical type I errors in the first panel of the second row of Figure \ref{fig:s1-CS}. This is because in Algorithm \ref{alm:cs}, if all the CS classifiers have empirical type I errors larger than $\alpha$, i.e.,  $r_{0,I}>\alpha$, then we choose $i^*=I$.} under $\alpha$.

\section{The NP classification paradigm}\label{sec:NP}

\subsection{An overview of the NP paradigm}
The NP classification paradigm is another statistical framework that addresses asymmetric priorities of type I and II errors \citep{scott2005neyman,scott2005comparison,rigollet2011neyman}.
Different from the CS paradigm, the NP paradigm aims to minimize the population type II error while controlling the population type I error below a pre-specified level $\alpha$: 
\begin{eqnarray}\label{eqn:np objective}
\phi_\alpha^* = \argmin_{\phi:R_0(\phi)\leq \alpha}R_1(\phi)\;.
\end{eqnarray}
We name $\phi_\alpha^*$ the (level-$\alpha$) NP oracle classifier.  Like the misclassification costs in the CS classification objective \eqref{eqn:cs objective}, the upper bound $\alpha$ in \eqref{eqn:np objective} reflects users' priorities for the type I error: a smaller $\alpha$ reflects a higher priority.


In practice, the NP oracle classifier is not achievable. Several algorithms were developed to construct data-dependent NP classifiers whose population type I error are under $\alpha$ with high probability \citep{zhao2016neyman,tong2018neyman, Tong.Xia.Wang.Feng.2020}. Especially, \cite{tong2018neyman} proposed an umbrella algorithm that adapts popular classification methods (e.g., support vector machine and random forest) to construct NP classifiers. The theoretical foundation of the umbrella algorithm is given by Proposition \ref{thm: nproc} in \cite{tong2018neyman}. For readers' convenience, we restate this proposition as follows.   

\begin{prop}{\citep{tong2018neyman}.}\label{thm: nproc}
Suppose that we divide the training data into two independent parts, one with observations from both classes $0$ and $1$ for training a classification method to obtain a scoring function $s: \mathcal{X}\rightarrow \bbR$,  and the other as a left-out class $0$ sample for choosing a threshold on classification scores. Applying  $s(\cdot)$ to the left-out class $0$ sample of size $m$, we denote the resulting classification scores as $T_1,\dots,T_m$, which are real-valued random variables. Then, we denote by $T_{(k)}$ the $k$-th order statistic. For a new observation $X$,  we can construct a classifier $\hat \phi_k (X) = \1(s(X)>T_{(k)})$. Then
\begin{eqnarray}
\bbP[R_0(\hat \phi_k) > \alpha] \leq \sum_{j=k}^{m}{m\choose j}(1-\alpha)^j\alpha^{m-j}\;.
\end{eqnarray}
That is, the probability that the population type I error of $\hat \phi_k$ exceeds $\alpha$ is under a constant that only depends on $n$, $k$, and $\alpha$. We call this probability the ``violation rate'' of $\hat \phi_k$ and denote its upper bound by $v(k)=\sum_{j=k}^{m}{m\choose j}(1-\alpha)^j\alpha^{m-j}$. When $T_i$'s are continuous, this bound is tight.
\end{prop}

According to Proposition \ref{thm: nproc}, to control the violation rate at a pre-specified $\delta$, we can construct an NP classifier $\hat\phi_{k^*}(\cdot) = \1(s(\cdot)>T_{(k^*)})$, where $k^* = \min\{k\in\{1,\dots,m\}:v(k)\leq \delta\}$. The minimum order is selected because among the classifiers that satisfy the type I error control, the one with the minimal type II error should be selected. The NP umbrella algorithm is summarized in Algorithm \ref{alm:np}.  

\begin{algorithm}[htb!]
\DontPrintSemicolon
\caption{\label{alm:np}
The NP umbrella algorithm}
\SetKw{KwBy}{by}
\SetKwInOut{Input}{Input}\SetKwInOut{Output}{Output}
\SetAlgoLined

\Input{$\mathcal{S}=\mathcal{S}^0\cup \mathcal{S}^1$: training data ($\mathcal{S}^0$ for class $0$ and $\mathcal{S}^1$ for class $1$) \\
$\alpha$: target upper bound on the type I error\\
$\delta$: target violation rate (i.e., the probability that the type I error exceeds $\alpha$)
}
$\mathcal{S}^0_1,\mathcal{S}^0_2 \leftarrow$ randomly split class $0$ data $\mathcal{S}^0$\;

$\mathcal{S}^0_2=\{(x_1,y_1),\dots,(x_m,y_m)\}\ (m = |\mathcal{S}^0_2|)$ \tcp*{left-out class 0 data}

$k^* = \min\left\{k\in\{1,\dots,m\}: \sum_{j=k}^{m}{m\choose j}(1-\alpha)^{j}\alpha^{m-j}\leq \delta\right\}$\;
$s=$ Classification-method($\mathcal{S}^0_1 \cup \mathcal{S}^1$)\\
\tcp*{train a classification method to obtain a scoring function}
$\mathcal{T}=\{T_1,\dots,T_m\}\leftarrow \{s(x_1),\dots, s(x_m)\}$\\
\tcp*{apply $s$ to $\mathcal{S}^0_2$ to obtain classification scores as candidate thresholds}
$\mathcal{T}^\text{sort}=\{T_{(1)},\dots,T_{(m)}\}$\\ 
\tcp*{sort candidate thresholds from the smallest to the largest}
\Output{$\hat\phi^{\textrm{NP}}(\cdot) = \1(s(\cdot)>T_{(k^*)})$ \tcp*{the NP classifier}}  
\end{algorithm}

\subsection{Population type I error in NP classification}

\begin{figure}[tbh!]
\centering
\includegraphics[width = .9\textwidth]{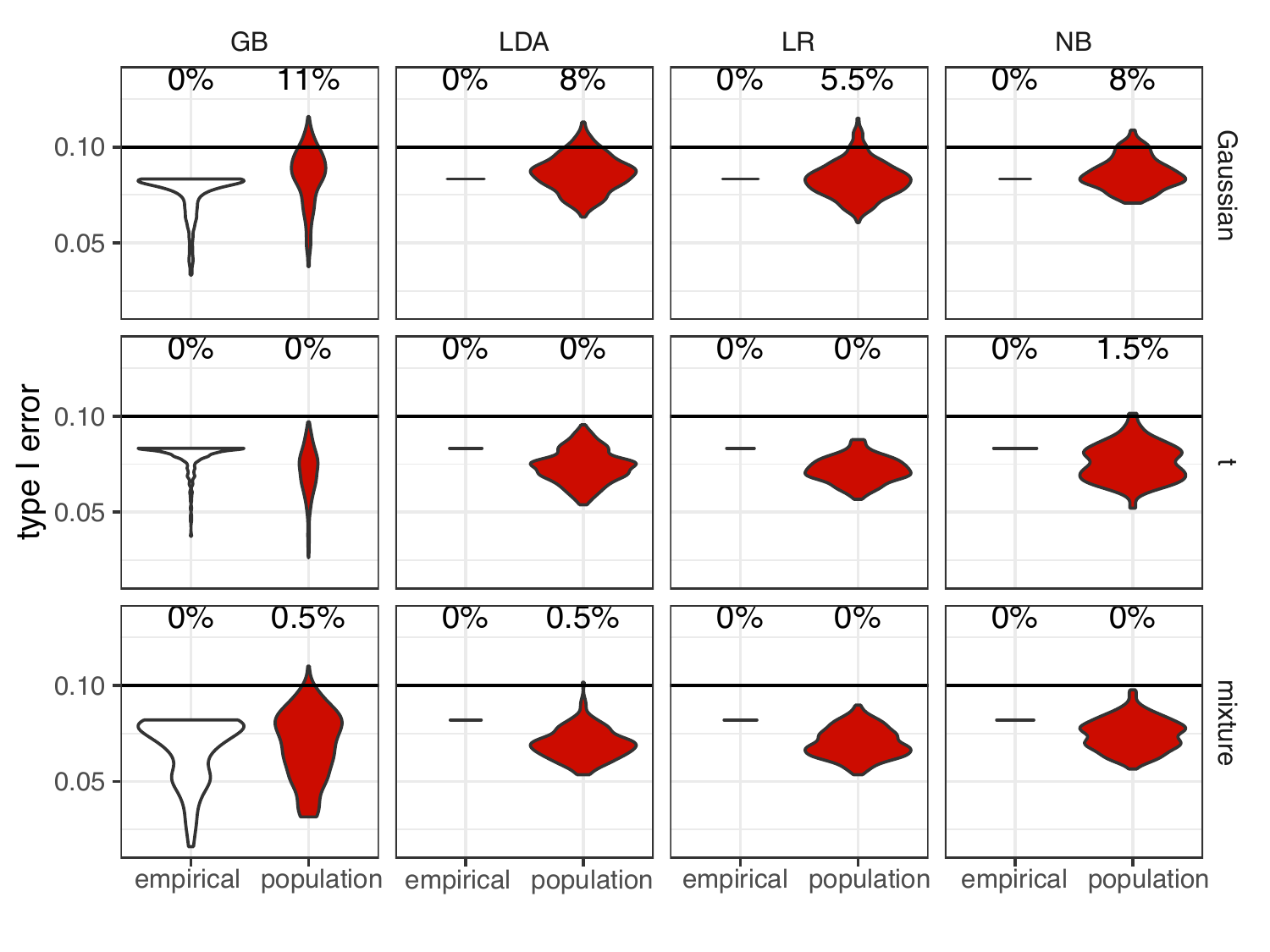}
\caption{\label{fig:s1-NP}
The empirical type I errors and population type I errors of NP classifiers trained by Algorithm \ref{alm:np}. The empirical type I error of each NP classifier is computed on the left-out class $0$ sample $\mathcal{S}_2^0$ in each simulation. We set the type I error upper bound as $\alpha = 0.1$ and use four classification methods: gradient boosting (GB), linear discriminant analysis (LDA), logistic regression (LR), and na\"ive Bayes (NB). The target violation rate is set as $\delta = 10\%$. We consider three data distributions: Gaussian, $t$, and Mixture. The violation rate of type I errors (the frequency of type I errors being greater than $\alpha=0.1$) in each case is summarized as a percentage and marked on top of each violin plot. 
}
\end{figure}

We repeat the simulation study in Section \ref{sec: cs-sim} by replacing CS classifiers with NP classifiers constructed by Algorithm \ref{alm:np}, and we numerically verify that the NP classifiers have their population type I errors under $\alpha$ with high probability (i.e., the violation rates are small).
In detail, from each of the three data distributions (Gaussian, Multivariate $t$, and Mixture), we simulate a training dataset with size $n_1 = 1{,}000$ and a large evaluation dataset with size $n_2 = 1{,}000{,}000$ to approximate the population. 
In each simulation, we use Algorithm \ref{alm:np} to train an NP classifier with $\alpha=0.1$ and $\delta = 0.1$, and we calculate the empirical type I error of the trained classifier using the left-out class $0$ data $\mathcal{S}^0_2$. Then we approximate the population type I error of the trained classifier using the large evaluation dataset. We repeat the simulation for $200$ times to calculate the violation rates of the empirical type I errors and the population type I errors.
Figure \ref{fig:s1-NP} summarizes the results and shows that the NP classifiers constructed by Algorithm \ref{alm:np} have population type I errors under $\alpha$ with probability at least $1-\delta$.

\section{From NP to CS: finding the type I error cost of an NP classifier}
\label{sec:exact-compare}
We have introduced CS and NP as two paradigms for prioritizing misclassification errors in asymmetric binary classification. It is evident that the CS objective \eqref{eqn:cs objective} and the NP objective \eqref{eqn:np objective} have a one-to-one correspondence at the population level.  However, it remains an unresolved question whether practical, data-dependent classifiers constructed from the same training data under the two paradigms would have a one-to-one correspondence. In this section, we establish an exact correspondence between an NP classifier and a CS classifier constructed by one of two CS implementations: the rebalancing and post-training approaches. Specifically, given an NP classifier, we can find the corresponding costs in these two CS implementations such that the resulting CS classifiers are just the NP classifier. Note that this correspondence is from NP to CS; that is, an NP classifier needs to be constructed first so that its corresponding costs can be found for the two CS implementations. 



\subsection{Theory}

We denote by $ f_0$ and $ f_1$ the conditional densities of $X|(Y=0)$ and $X|(Y=1)$, respectively. Recall that $\pi_0 = \bbP(Y=0)$ and $\pi_1 = \bbP(Y=1)$. Then the regression function, which is the scoring function of the Bayes classifier, can be expressed as 

\begin{eqnarray}
  \eta(X) = \bbP(Y=1|X)=\frac{ f_1(X)\pi_1}{ f_0(X)\pi_0+ f_1(X)\pi_1}\;.
\end{eqnarray}
At the sample level, as in Algorithm \ref{alm:np}, we divide the training data $\mathcal{S}$ into two subsets: $\mathcal{S}^0_1 \cup \mathcal{S}^1$: a mixture of class $0$ and class $1$ observations used to estimate $\eta(\cdot)$; $\mathcal{S}^0_2$: a left-out class $0$ sample used to select the threshold in an NP classifier.
In this context, we can estimate the scoring function $\eta(\cdot)$ from $\mathcal{S}^0_1 \cup \mathcal{S}^1$ by
\begin{eqnarray}\label{eqn:hat eta}
  \hat \eta(X) =\frac{\hat f_1(X)\hat\pi_1}{\hat f_0(X)\hat\pi_0+\hat f_1(X)\hat\pi_1}\;,
\end{eqnarray}
where $\hat\pi_0$ and $\hat\pi_1$ are respective estimates of $\pi_0$ and $\pi_1$, and  $\hat f_0(\cdot)$  and $\hat f_1(\cdot)$ are respectively estimates of $f_0(\cdot)$ and $f_1(\cdot)$. Then a plug-in version of the Bayes classifier is $\hat\phi_\text{classic}(X) = \1(\hat \eta(X)> 1/2)$, where $\hat\eta(\cdot)$ is defined in \eqref{eqn:hat eta}.  

Now we construct an NP classifier using $s(\cdot)=\hat \eta(\cdot)$. Let $t_\text{NP}=T_{(k^*)}$, which is found from $\mathcal{S}^0_2$ by Algorithm \ref{alm:np}. An NP classifier can then be constructed as 
$\hat\phi_\text{NP}(X) = \1(\hat \eta(X)> t_\text{NP})$. Next, we will show how to decide the misclassification costs corresponding to the NP classifier in two CS implementations: rebalancing and post-training.

\subsubsection{Special case 1: the rebalancing approach}
Recall the CS objective function
\begin{eqnarray*}
c_0R_0(\phi_\text{CS}) + c_1R_1(\phi_\text{CS})\;.
\end{eqnarray*}
In the rebalancing approach, the class priors are set to the misclassification costs $c_0$ and $c_1$ instead of the two class proportions $\hat\pi_0$ and $\hat\pi_1$ in the training data. As such, the scoring function becomes
\begin{eqnarray}\label{eqn: rebalancing}
\tilde \eta(X) =\frac{\hat f_1(X)c_1}{\hat f_0(X)c_0+\hat f_1(X)c_1}\;.
\end{eqnarray}
Then the CS classifier that mimics the Bayes classifier takes the form 
$\hat\phi_\text{CS}^{\text{r}}(X) = \1(\tilde \eta(X)> 1/2)$ \citep{xia2020intentional}.
Proposition \ref{thm: rebalance} shows that, given $\hat\phi_\text{NP}$,  it is possible to construct an equivalent $\hat\phi_\text{CS}^{\text{r}}$. 

\begin{prop}\label{thm: rebalance}
Given a level-$\alpha$ NP classifier $\hat\phi_\emph{NP}(X) = \1(\hat \eta(X)> t_\emph{NP})$, where $\hat\eta(\cdot)$ is defined in \eqref{eqn:hat eta} and $0\le t_\emph{NP} \le 1$, if we assign a type I error cost 
\begin{eqnarray}
  c_0=c_0(t_\emph{NP}, \hat\pi_0)=\frac{t_\emph{NP}\hat\pi_0}{(1-t_\emph{NP})(1-\hat\pi_0) + t_\emph{NP}\hat\pi_0}\;,
\end{eqnarray}
then the resulting CS classifier $\hat \phi_\emph{CS}^{\emph{r}}(X) = \1(\tilde \eta(X)>1/2) = \hat\phi_\emph{NP} (X)$, where $\tilde\eta(\cdot)$ is defined in \eqref{eqn: rebalancing}. That is, $R_0(\hat \phi_\emph{CS}^{\emph{r}})\leq \alpha$ with high probability.
\end{prop}

\subsubsection{Special case 2: the post-training approach }

The post-training approach constructs a CS classifier as
\begin{align}
\begin{split}
\hat\phi_\text{CS}^{\text{pt}}(X) 
&= 
\1\left(c_1\hat \eta(X) >c_0 (1-\hat\eta(X))\right)
\\
&= \1\left(\frac{\hat\eta(X)}{1-\hat\eta(X)} > \frac{c_0}{c_1}\right)\\
&=\1\left(\hat\eta(X)>c_0\right)\;,
\end{split}
\end{align}
where the last equality holds because $c_1 = 1-c_0$.
Proposition \ref{thm: post}, which is self-evident, shows that, given $\hat\phi_\text{NP}$,  we can construct an equivalent $\hat\phi_\text{CS}^{\text{pt}}$.

\begin{prop}\label{thm: post}
Given a level-$\alpha$ NP classifier $\hat\phi_\emph{NP}(X) = \1(\hat \eta(X)> t_\emph{NP})$, where the scoring function $\hat\eta(X)=\widehat\bbP(Y=1|X)$ is the estimated posterior probability either defined in \eqref{eqn:hat eta} or constructed by other classification methods (e.g., logistic regression and random forest), if we assign a type I error cost $c_0=t_\emph{NP}$, 
then the resulting CS classifier $\hat\phi_\emph{CS}^{\emph{pt}}(X) = \1(\hat \eta(X)>c_0) = \hat\phi_\emph{NP}(X)$. That is,  $R_0(\hat \phi_\emph{CS}^{\emph{pt}})\leq \alpha$ with high probability.
\end{prop}

\subsection{Simulation results}\label{sec:rebalance-sim}

We perform a sanity check of Propositions \ref{thm: rebalance} and \ref{thm: post} using simulation studies. 
For the rebalancing CS approach, we implement it with three classification methods: linear discriminant analysis, quadratic discriminant analysis, and na\"ive Bayes. 
For the post-training CS approach, we implement it with logistic regression as an example.

\begin{figure}[tbh]
\includegraphics[width = \textwidth]{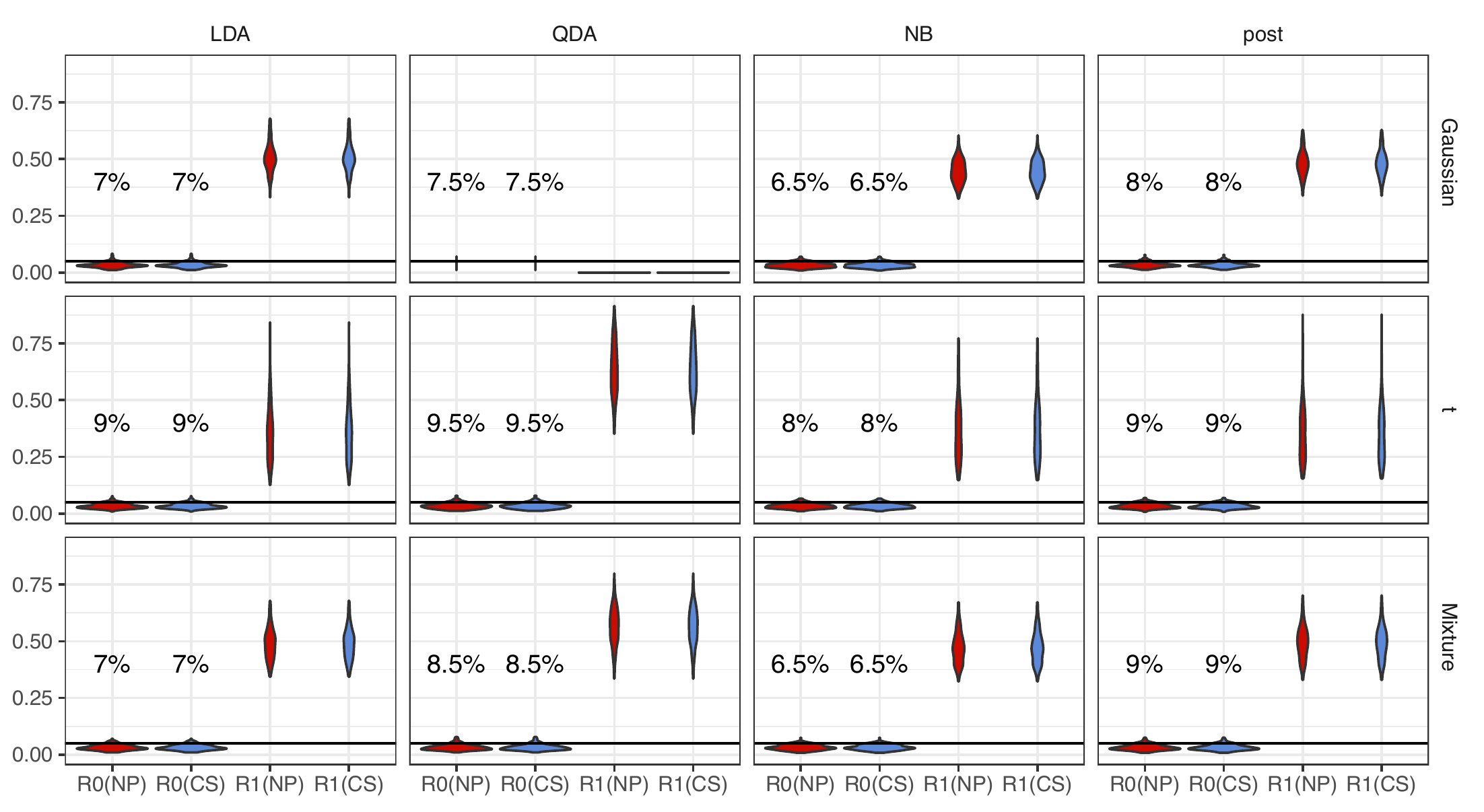}
\caption{\label{fig:rebalance}
The population type I errors ($R_0$) and type II errors ($R_1$) of NP classifiers and their corresponding rebalancing or post-training CS classifiers. We use the rebalancing approach with three classification methods: linear discriminant analysis (LDA), quadratic discriminant analysis (QDA), and na\"ive Bayes (NB), corresponding to the first three columns. We use the post-training approach with logistic regression (the fourth column). The target violation rate is set as $\delta = 10\%$. We consider three data distributions: Gaussian, $t$, and Mixture. The violation rate of the population type I errors (i.e., the frequency of population type I errors exceeding $\alpha = 0.05$) in each case is summarized as a percentage and marked on top of each violin plot.
}
\end{figure}

From each of the three data distributions (Gaussian, Multivariate $t$, and Mixture) in Section \ref{sec: cs-sim}, we simulate a training dataset with size $n_1 = 1{,}000$ and a large evaluation dataset with size $n_2 = 1{,}000{,}000$. We split the training data into two subsets: $\mathcal{S}^0_1 \cup \mathcal{S}^1$: a mixture of classes $0$ and $1$ for obtaining the scoring function $\hat \eta(\cdot)$; $\mathcal{S}^0_2$: a left-out class $0$ sample for selecting the threshold $t_\text{NP}=T_{(k^*)}$ by Algorithm \ref{alm:np}. To construct an NP classifier, we set the left-out class $0$ sample size $m = |\mathcal{S}_2^0| = 200$, the target type I error upper bound $\alpha = 0.05$, and the target violation rate $\delta = 0.1$. Based on the NP classifier, we construct the corresponding rebalancing or post-training CS classifier based on Proposition \ref{thm: rebalance} or \ref{thm: post}, respectively. 
Then we approximate the population type I and type II errors of these NP and CS classifiers using the large evaluation dataset. We repeated the simulation for $200$ times. Figure \ref{fig:rebalance} indicates that the NP classifiers and their corresponding rebalancing or post-training CS classifiers have identical population type I and type II errors, confirming each pair of NP and CS classifiers' equivalence.  

\section{From CS to NP: estimating an upper bound of the population type I error of a CS classifier}
\label{sec:gen-alg}
Except for the rebalancing and post-training CS implementations discussed in Section \ref{sec:exact-compare}, in general, it is infeasible to construct a CS classifier that is the same as a given NP classifier.  
The other direction is to evaluate a given CS classifier's population type I error performance.  We hold that it is impossible to specify $(\alpha, \delta)$, the target population type I error upper bound and the target violation rate, to train an NP classifier to be the same as the given CS classifier.  However, a less ambitious goal is possible. Given a CS classifier, by considering it as fixed, we can estimate an upper bound on its population type I error using an independent class $0$ sample that is not used for training. Our goal is to have the estimated upper bound (a random variable) above the classifier's unknown population type I error (a fixed value) with high probability. We believe that estimating the population type I error upper bound will assist the interpretation and assessment of a CS classifier in practice.

In this section, we propose two algorithms to achieve this goal. We first introduce a theoretical result that motivates the design of the algorithms. Then we introduce the algorithms and present their numerical results.

\subsection{Theoretical Results}

Given a fixed upper bound $\alpha$ on the population type I error, the next proposition shows an upper bound on the ``violation rate'' of a surrogate classifier---which is constructed from a left-out class $0$ sample of size $m$---of a fixed CS classifier. Note that the left-out class $0$ sample is not used to train the CS classifier, and the ``surrogate violation rate'' is defined as the probability (over all possible left-out class $0$ samples of size $m$) that the population type I error of the surrogate classifier is greater than $\alpha$.

\begin{prop}\label{thm: approx-bound}
For a given CS classifier $\hat \phi_\emph{CS}(\cdot) = \1(s(\cdot)>t_\emph{CS})$, we consider $s(\cdot)$ and $t_\emph{CS}$ as fixed. Denote by $F$ the cumulative distribution function of $s(X^0)$, where $X^0\sim X|(Y=0)$.  We apply the scoring function $s(\cdot)$ to a left-out class $0$ sample of size $m$.   Denote the resulting classification scores by $T_1,\dots,T_m$, among which $T_{(k)}$ denotes the $k$-th order statistic. 
If $T_{(1)}\leq t_\emph{CS}$, we consider the surrogate order 
$k^*_s = \max\{k\in\{1,\dots,m\}: T_{(k)}\leq t_\emph{CS}\}$
and construct a surrogate classifier $\hat \phi_{k^*_s}(\cdot) = \1(s(\cdot)>T_{(k^*_s)})$.
Then $R_0(\hat \phi_{k^*_s}) \geq R_0(\hat \phi_\emph{CS})$, and 
\begin{align}
\bbP_m(R_0(\hat \phi_{k^*_s})&>\alpha)
\begin{cases}
=1 & \text{if}\ t_\emph{CS} < F^{-1}(1-\alpha)\\
\leq (2-\alpha-F(t_\text{CS}))^m-(1-F(t_\text{CS}))^m & \text{if}\ t_\emph{CS} \geq F^{-1}(1-\alpha)
\end{cases}
\;,
\end{align}
where $\bbP_m$ is with respect to the randomness induced by the left-out class $0$ sample of size $m$. We denote $\delta_s=(2-\alpha-F(t_\text{CS}))^m-(1-F(t_\text{CS}))^m$.
\end{prop}

Note that $\bbP_m(R_0(\hat \phi_{k^*_s})>\alpha)$ in Proposition \ref{thm: approx-bound} is not the violation rate defined in the previous sections for two reasons: the classifier is the surrogate classifier, not the actual CS classifier; the randomness lies in a left-out class $0$ sample not used for training, while the CS classifier is fixed.  Nevertheless, Proposition \ref{thm: approx-bound} helps motivate algorithms to estimate an upper bound of $R_0(\hat \phi_{\text{CS}})$. Concretely, the population type I error of the surrogate classifier, $R_0(\hat \phi_{k^*_s})$, is always greater than $R_0(\hat \phi_\text{CS})$ and can be controlled under $\alpha$ with a surrogate violation rate no greater than $\delta_s=(2-\alpha-F(t_\text{CS}))^m-(1-F(t_\text{CS}))^m$ if $t_\text{CS} \geq F^{-1}(1-\alpha)$. 

Note that since the CS classifier is fixed, $F(t_{\text{CS}})$ and $\delta_s$ are fixed.\footnote{Note that when we also consider the randomness of the CS classifier,  $\delta_s$ would not be a fixed number but rather depends on the random quantity $F(t_{\text{CS}})$, in which $F(\cdot)$ inherits randomness from $s(\cdot)$, and $t_{\text{CS}}$ is possibly random too.} Then given a pre-specified $\delta_s=\delta$, we can solve for $\alpha$ as a function of $F(t_{\text{CS}})$. However, $F(\cdot)$ is unobservable, and we need to estimate $F(t_{\text{CS}})$. With estimates of $F(t_{\text{CS}})$, we will have the corresponding estimates of $\alpha$, which would allow us to construct an estimated upper bound $\hat\alpha$ (random) of $R_0(\hat \phi_\emph{CS})$ (fixed) such that $\hat\alpha > R_0(\hat \phi_\emph{CS})$ with high probability.



\subsection{The TUBE Algorithms}

We propose two algorithms, TUBEc  and TUBE (estimating the \textbf{T}ype I error \textbf{U}pper \textbf{B}ound of a cost-sensitive classifi\textbf{E}r), both of which are motivated by Proposition \ref{thm: approx-bound}. 
The first algorithm, TUBEc (TUBE-core), contains the core component to estimate a given CS classifier's type I error upper bound, and it depends on the availability of a left-out class $0$ sample.
However, in practice it is often desirable to construct a classifier using all the available data, especially when the available sample size is small. In view of this, we further propose a second algorithm, TUBE, which is built upon TUBEc and aims to estimate a high-probability type I error upper bound of a CS classifier trained on all the available data. 
We will introduce TUBEc first and TUBE next.

\subsubsection{The TUBEc Algorithm}
\begin{algorithm}[h]
\DontPrintSemicolon
\caption{The TUBEc algorithm for estimating type I error upper bound\label{alm:tube}}
\SetKw{KwBy}{by}
\SetKwInOut{Input}{Input}
\SetKwInOut{Output}{Output}
\SetAlgoLined
\Input{$\1(s(\cdot)> t_{\text{CS}})$: CS classifier \\ 

$\mathcal{S}^0_2 = \{x_1,\dots,x_m\}$: left-out class 0 sample\\
$\delta$: pre-specified violation rate
}

$\mathcal{T}=\{T_1,\dots,T_m\}\leftarrow \{s(x_1),\dots,s(x_m)\}$\;
\For{$b\gets1$ \KwTo $B$}{
    $\mathcal{T}^b = \{T^b_{1},\dots,T^b_{m}\}$ $\leftarrow$ {sample($\mathcal{T}$, size  = $m$, replace = true)}
    $\{T^b_{(1)},\dots,T^b_{(m)}\}\leftarrow\text{sort}(\mathcal{T}^b)$\;
     \uIf{$T^b_{(1)}>t_\emph{CS}$}{
    $\hat\alpha_b = 1$\;
     }
    \Else{
    $k_{s,b}^* = \max\{k\in\{1,\dots,m\}: T^b_{(k)}\leq t_\text{CS}\}$\;
    $\widehat{F}_b(t_\text{CS}) = k_{s,b}^*/m$\;
	$\hat\alpha_b = 2 - \widehat{F}_b(t_\text{CS}) - \left[\delta+\left(1-\widehat{F}_b(t_\text{CS})\right)^m\right]^\frac{1}{m}$\;
    \tcp*{solve for $\alpha$ based on Proposition 4}
  }
}
\Output{estimated type I error upper bound \\$\hat\alpha=(1-\delta)$-th quantile of $\{\hat\alpha_b\}_{b=1}^B$\tcp*{the TUBEc estimator}}
\end{algorithm}


The TUBEc algorithm (Algorithm \ref{alm:tube}) estimates $F(\cdot)$ on the left-out class $0$ sample using the nonparametric bootstrap procedure. 
TUBEc then estimates the type I error upper bound of a CS classifier based on Proposition \ref{thm: approx-bound}. TUBEc is a generic algorithm applicable to all CS classifiers, which may be constructed by any of the pre-training, in-training, and post-training approaches (Table~\ref{tab:cs-summary}). 
Specifically, for a given classification scoring function $s(\cdot)$, a pre-determined classification score threshold $t_\text{CS}$, and a left-out class 0 sample $\mathcal{S}^0_2$, TUBEc estimates the type I error upper bound of the CS classifier $\1( s(\cdot)> t_{\text{CS}})$.
In each iteration $b$, TUBEc first obtains a bootstrap sample of the classification scores on the left-out class 0 sample, and then it estimates a type I error upper bound $\hat\alpha_b$. After $B$ iterations, TUBEc calculates the final estimated type I error upper bound as $\hat\alpha=(1-\delta)$-th quantile of $\{\hat\alpha_b\}_{b=1}^B$.

We apply the TUBEc algorithm to multiple simulation settings to estimate the type I error upper bound of CS classifiers. We consider two classification methods, logistic regression and gradient boosting, combined with the stratification approach for constructing CS classifiers. For each classification method, we consider three left-out sample sizes ($|\mathcal{S}^0_2|=m$): $50, 100,\ \text{and}\ 200$, and we set the type I error cost $c_0$ to $0.7$. The simulated data are drawn from the three distributions used in previous simulations, Gaussian, Multivariate $t$, and Mixture (Section \ref{sec: cs-sim}). We set the desired violation rate to $\delta = 0.1$ for all settings.  In each setting, we first train a CS classifier on a training sample of size $500$ and then apply the TUBEc algorithm to $100$ independent left-out class 0 samples to estimate the type I error upper bound of the CS classifier. We repeat this simulation experiment three times for every combination of a classification method, a left-out sample size, and a data distribution. 

\begin{figure}[tbhp!]
\centering
\includegraphics[width=.95\textwidth]{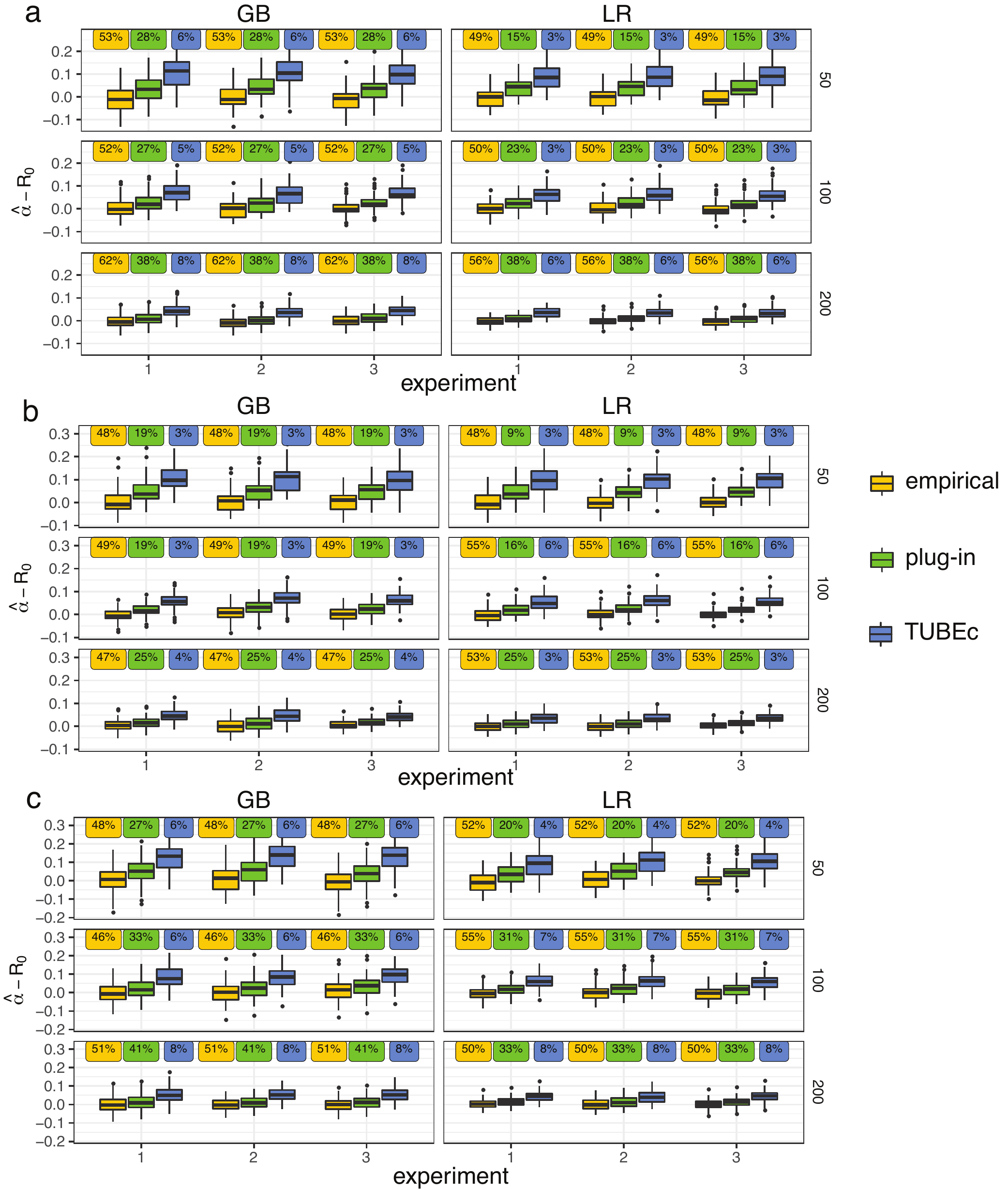}
\caption{Performance of the TUBEc estimator ($\hat\alpha^\text{TUBEc}$), the empirical estimator ($\hat\alpha^\text{emp}$), and the plug-in estimator ($\hat\alpha^\text{plug-in}$). 
The simulated data are drawn from three distributions: (a) Gaussian, (b) Multivariate $t$, and (c) Mixture. We use the stratification approach with two classification methods, gradient boosting (GB) and logistic regression (LR), to construct CS classifiers.
The boxplots show the distributions of the difference between the estimated type I error upper bound ($\hat\alpha$) and the actual population type I error of the given CS classifier ($R_0$) (approximated on a large sample of size $10^6$).
The violate rate (percentage of simulations in which $R_0 > \hat\alpha$) is labeled on the top of each boxplot. 
The left-out class 0 sample size is labeled on the right of each row. The horizontal axis indicates the three experiments as replications of each simulation setting.
\label{fig:TUBEc}
}
\end{figure}

We compare the TUBEc estimator with the empirical estimator $\hat\alpha^{\text{emp}}$ defined as the empirical type I error of the CS classifier on the left-out sample. Moreover, to demonstrate the necessity of the bootstrap procedure in the TUBEc algorithm, we consider a plug-in estimator defined directly based on Proposition \ref{thm: approx-bound}. Concretely, instead of using estimates $\hat\alpha_1, \ldots, \hat\alpha_B$ based on the bootstrapped class $0$ scores $\mathcal{T}^1, \ldots, \mathcal{T}^B$, we define the plug-in estimator $\hat\alpha^{\text{plug-in}} = 2 - \widehat{F}(t_\text{CS}) - [\delta+(1-\widehat{F}(t_\text{CS}))^m]^{1/m}$, where $\widehat{F}(t_\text{CS})$ is estimated from $\mathcal{T}$, the left-out class $0$ scores. Figure \ref{fig:TUBEc} shows the distribution of the difference between each estimator and the population type I error of the given CS classifier under each simulation setting. We expect that the distributions corresponding to a good estimator should have probabilities close to $\delta = 10\%$ for negative differences, indicating that the estimator is greater than the population type I error with high probability. Among the three estimators (empirical, plug-in, and TUBEc), only the TUBEc estimator achieves this property.


\subsubsection{The TUBE algorithm}

The TUBEc algorithm estimates a type I error upper bound for a CS classifier whose training sample does not include the left-out class $0$ sample. However, in real applications, it is often desirable or even necessary to train a CS classifier using the whole sample $\mathcal{S}$, since a larger training sample size in general leads to better classification performance. 
Therefore, we further propose TUBE (Algorithm~\ref{alm:tube+}) as a modified version of the TUBEc algorithm to estimate the type I error upper bound of a CS classifier constructed using $\mathcal{S}$. The central strategy is to adjust up the empirical type I error computed on $\mathcal{S}$ by the difference between the TUBEc estimate and this empirical type I error. Note that the TUBEc estimate is only calculable when a separate left-out class $0$ sample is available; however, as we have used up all the available observations, we cannot calculate this difference but have to mimic it by an average of the differences evaluated by random partitioning of $\mathcal{S}$, in which each partition leads to (1) a mixed class sample for training a CS classifier and calculating the empirical type I error and (2) a left-out class $0$ sample for calculating the TUBEc estimate.


\begin{algorithm}[htbp]
\DontPrintSemicolon
\caption{The TUBE algorithm for estimating type I error upper bound\label{alm:tube+}}
\SetKw{KwBy}{by}
\SetKwInOut{Input}{Input}
\SetKwInOut{Output}{Output}
\SetAlgoLined
\Input{$\mathcal{S}=\mathcal{S}^0\cup \mathcal{S}^1$: training sample ($\mathcal{S}^0$ for class $0$ and $\mathcal{S}^1$ for class $1$) \\
$c_0$: type I error cost\\
$\delta$: pre-specified violation rate
}
$\1( s(\cdot) > t_{\text{CS}})=$CS-classifier($\mathcal{S}$, $c_0)$\\
$\tilde\alpha =$ empirical type I error of $\1(s(\cdot)>t_\text{CS})$ on $\mathcal{S}$
\tcp*{on training sample}

\For{$b_1\gets1$ \KwTo $B_1$}{
  $\mathcal{S}^0_{1b_1}, \mathcal{S}^0_{2b_1}\leftarrow$ randomly split class $0$ sample $\mathcal{S}^0$\;

  $\mathcal{S}_{b_1} = \mathcal{S}^1 \cup \mathcal{S}^0_{1b_1}$
  \tcp*{training data for CS classifier} 
  $\1(s_{b_1}(\cdot)> t_{\text{CS}, b_1})=$ CS-classifier($\mathcal{S}_{b_1}$, $c_0$) \;
  $\hat{\alpha}^\text{TUBEc}_{b_1}=$TUBEc($\1(s_{b_1}(\cdot)> t_{\text{CS}, b_1})$, $\mathcal{S}^0_{2b_1}$, $\delta$)\tcp*{the TUBEc estimator}

  $\tilde\alpha_{b_1}=$ empirical type I error of $\1(s_{b_1}(x)>t_{\text{CS},b_1})$ on $\mathcal{S}_{b_1}$ 
}
  
  $\hat{\alpha}^\text{TUBE} = \tilde \alpha + \frac{1}{B_1}\sum_{b_1=1}^{B_1} (\hat \alpha^{\text{TUBEc}}_{b_1} - \tilde \alpha_{b_1})$\tcp*{the TUBE estimator}

\Output{$\hat{\alpha}^\text{TUBE}$}
\end{algorithm}

%

\begin{figure}[tbh!]
\centering
\includegraphics[width=\textwidth]{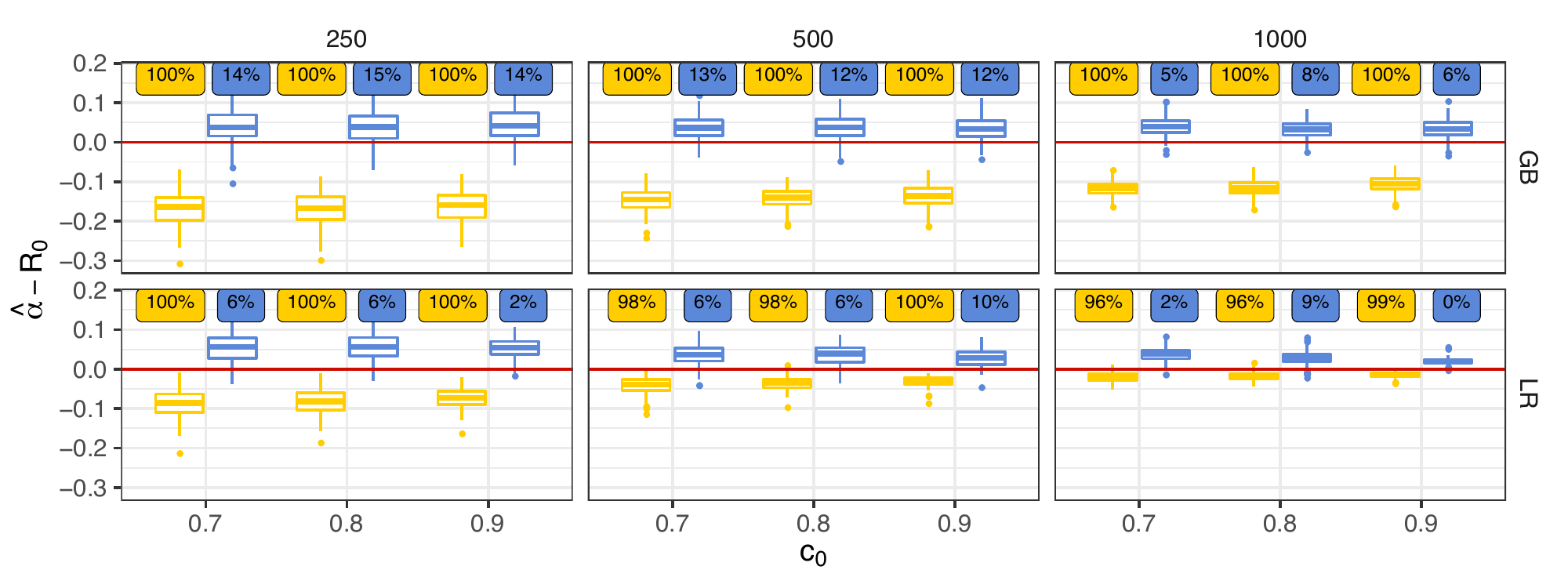}
\caption{
Performance of the TUBE estimator ($\hat\alpha^\text{TUBE}$; blue) and the empirical estimator ($\hat\alpha^\text{emp}$; yellow). Data are generated from the Gaussian distribution (Section~\ref{sec: cs-sim}). We use the stratification approach with two classification methods, gradient boosting (GB) and logistic regression (LR), to construct CS classifiers.
The boxplots show the distributions of the difference between the estimated type I error upper bound ($\hat\alpha$) and the actual population type I error of the given CS classifier ($R_0$) (approximated on a large sample of size $10^6$).
The horizontal axis denotes the type I error cost ($c_0 = 0.7$, $0.8$, or $0.9$). The training sample size ($250$, $500$, or $1000$) is labeled on the top of each column.
The violation rate (percentage of simulations in which $R_0 > \hat\alpha$) is labeled on the top of each boxplot.
\label{fig:tube}
}
\end{figure}

In Figure \ref{fig:tube}, using training samples generated from the Gaussian distribution (Section~\ref{sec: cs-sim}), we construct CS classifiers using the stratification approach with two classification methods: gradient boosting and logistic regression. Given these CS classifiers, we compare the TUBE estimator $\hat\alpha^\text{TUBE}$ (with $\delta = .1$) with the empirical estimator $\hat\alpha^\text{emp}$, which is defined as the empirical type I error of a CS classifier on the training sample. (Results for the Multivariate $t$ and Mixture distributions are in Figure   \ref{fig:tube-supp} in the Appendix).  
Granted, $\hat\alpha^\text{TUBE}$ is not guaranteed to dominate the true type I error of a CS classifier with at least $1-\delta$ probability. Still, Figure \ref{fig:tube} shows that $\hat\alpha^\text{TUBE}$ is a high-probability upper bound on the true type I error. Empirically, we observe that the actual violation rate (i.e., the probability that $\hat\alpha^\text{TUBE}$ is below the true type I error) is below $1.5 \delta$ in all cases, and it falls under $\delta$ when the training sample size is large. In contrast, $\hat\alpha^\text{emp}$ fails to be a high-probability upper bound, as expected.   

\section{TUBE-assisted CS classification for type I error control}\label{sec:simulation}


In this section, we introduce a TUBE-assisted algorithm for selecting the type I error cost $c_0$ in CS classification so that the resulting CS classifier has its population type I error under a target upper bound $\alpha$ with high probability. Note that the TUBE algorithm (Algorithm \ref{alm:tube+}) provides an estimated high-probability upper bound on the population type I errors of CS classifiers constructed with a given cost $c_0$. Hence, if we have a target type I error upper bound $\alpha$, we can supply candidate type I error costs into the TUBE algorithm and evaluate their corresponding TUBE estimates; then, we can compare these TUBE estimates with $\alpha$ and pick the smallest type I error cost whose TUBE estimate is under $\alpha$. We refer to this TUBE-assisted algorithm by TUBE-CS and describe it in Algorithm \ref{alm:tube-cs}.

To evaluate TUBE-CS, we compare it with three algorithms: the TUBEc-assisted algorithm (TUBEc-CS; Algorithm \ref{alm:TUBEc-cs} in Appendix), the vanilla CS implementation (Algorithm \ref{alm:cs}), and the NP umbrella algorithm (Algorithm \ref{alm:np}). Note that TUBEc-CS is similar to TUBE-CS in that it replaces TUBE estimates with TUBEc estimates. Since TUBE-c requires a left-out class $0$ sample, TUBEc-CS requires sample splitting, unlike TUBE-CS. We have shown that the vanilla CS implementation cannot construct CS classifiers with population type I errors under $\alpha$ with high probability (Section~\ref{sec: cs-sim}), and here we use it as a negative control. The NP umbrella algorithm is designed exactly for this type I error control purpose; however, it requires sample splitting for constructing NP classifiers. All the four algorithms---TUBE-CS, TUBEc-CS, vanilla-CS, and NP---construct classifiers given a classification method, a target type I error upper bound $\alpha$, and a pre-specified violation rate $\delta$. The goal is to evaluate whether their constructed classifiers have population type I errors under $\alpha$ with a high probability close to $1-\delta$.


\begin{algorithm}[htb!]
\DontPrintSemicolon
\caption{\label{alm:tube-cs}
TUBE-assisted CS implementation (TUBE-CS)}
\SetKw{KwBy}{by}
\SetKwInOut{Input}{Input}\SetKwInOut{Output}{Output}
\SetAlgoLined

\Input{$\mathcal{S}=\mathcal{S}^0\cup \mathcal{S}^1$: training data ($\mathcal{S}^0$ for class $0$ and $\mathcal{S}^1$ for class $1$)\\
$\alpha$: target upper bound on the type I error\\
$c_{0,1} < c_{0,2} < \dots < c_{0,I}$: candidate type I error costs\\
$\delta$: pre-specified violation rate
}

\For{$i\gets1$ \KwTo $I$}{
  $\hat\alpha_i$ = TUBE($\mathcal{S}, c_{0,i}, \delta$)
}
$i^* = \min\{i: \hat\alpha_i\leq \alpha\}$\;

\Output{$\hat{\phi}^{\textrm{TUBE-CS}}(\cdot) =$ CS-classifier$(\mathcal{S}, c_{0,i^*})$\tcp*{the TUBE-CS classifier}}
\end{algorithm}

\begin{figure}[tbh!]
\centering
\includegraphics[width=\textwidth]{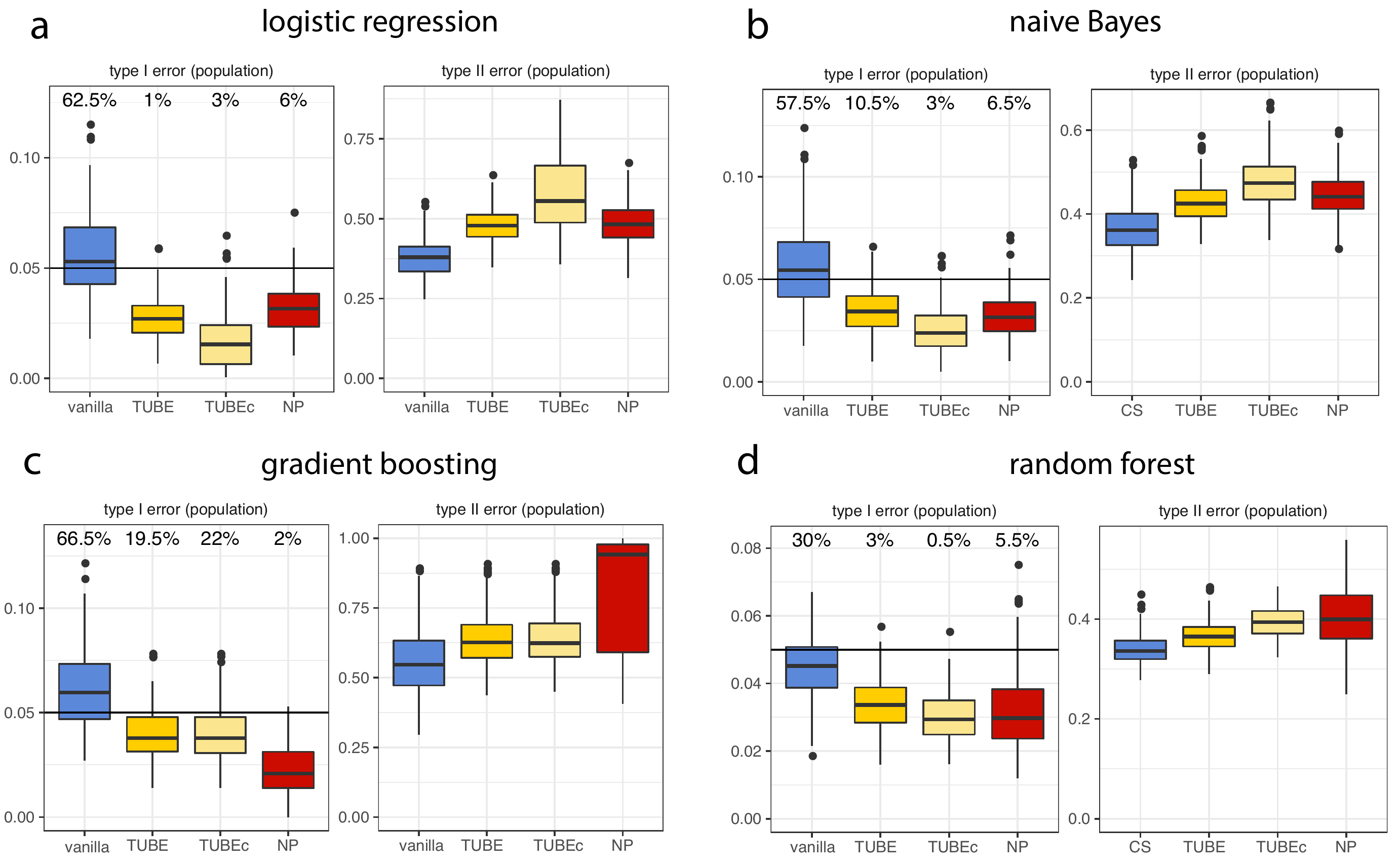}
\caption{
Comparison of four algorithms: vanilla-CS (``vanilla''), TUBE-CS (``TUBE''), TUBEc-CS (``TUBEc''), and NP umbrella (``NP'') algorithms. The population type I and II errors (approximated on the large evaluation sample) are shown as boxplots for four classification methods: (a) logistic regression, (b) na\"ive Bayes, (c) gradient boosting, and (d) random forest. The violation rate (percentage of simulations in which the population type I error exceeds $\alpha=0.05$) is labeled on the top of each boxplot.
\label{fig:s5}
}
\end{figure}

To compare the four algorithms, we design a simulation study with $\alpha = 0.05$ and $\delta = 0.1$. 
We pair each algorithm with four classification methods: logistic regression, na\"ive Bayes, gradient boosting, and random forest. 
We simulate data from the Gaussian distribution (Section \ref{sec: cs-sim}) with $d=30$. In each simulation, we generate a training sample with size $n_1 = 1,000$ and a large evaluation sample with size $n_2 = 1,000,000$ (to approximate the population). We use the stratification approach to construct CS classifiers.

Figure \ref{fig:s5} shows the comparison results. As expected, the NP umbrella algorithm is the only one whose constructed classifiers have population type I errors under $\alpha$ with more than $1-\delta$ probability in every setting. The TUBE-CS and TUBEc-CS algorithms have achieved desirable type I error control with logistic regression, na\"ive Bayes, and random forest, but not with gradient boosting. The vanilla-CS algorithm cannot construct classifiers with desirable type I error control, regardless of the classification method. 

Regarding the type II error, classifiers constructed by the TUBE-CS algorithm generally have smaller population type II errors than those of the classifiers constructed by the NP umbrella algorithm. This advantage comes from the fact that the TUBE-CS algorithm trains scoring functions in classifiers using all the training data, while the NP umbrella algorithm leaves out a portion of the class $0$ sample from the training of scoring functions. Unlike the TUBE-CS algorithm, the TUBEc-CS algorithm tends to be too conservative in controlling the type I error and thus does not outperform the NP umbrella algorithm in terms of the type II error. Together, these results suggest that the NP umbrella algorithm should always be preferred if a strict high-probability control on the type I error is desired. However, when such a strict control can be loosened a bit and the training sample size is moderate or small, the TUBE-CS algorithm is preferred because it delivers smaller type II errors thanks to better scoring functions trained on all available data.

\section{Comparison of vanilla-CS and TUBE-CS algorithms on real datasets}\label{sec:realdata}

In this section, we compare the vanilla-CS and TUBE-CS algorithms on four real datasets from the UCI Machine Learning Repository \citep{Dua:2019}.
A summary of these four datasets is in Table \ref{tab:s6-data}.
The \texttt{Diabetes} dataset (generated by the National Institute of Diabetes and Digestive and Kidney Diseases) allows the training of a classifier to predict whether a patient has diabetes from his/her diagnostic measurements; the more severe type of classification error is the misprediction of a diabetes patient as undiseased.
The \texttt{Thyroid} dataset (generated by the Garavan Institute) allows the training of a classifier to predict whether a subject has the euthyroid sick syndrome; the more severe type of classification error is the misprediction of a euthyroid sick patient as undiseased.
The \texttt{Breast cancer} dataset allows the training of a classifier to predict the state (benign or malignant) of a mammographic mass lesion, so as to help physicians decide whether to perform a biopsy on the lesion or just schedule a short-term follow-up examination; the more severe type of classification error is the misprediction of a malignant lesion as benign.
The \texttt{Marketing} dataset of a Portuguese banking institution allows the training of a classifier to predict if a client will subscribe (yes or no) to a term deposit based on the information gathered in a phone call; the more severe type of classification error is missing a potential subscriber.

\begin{table}[tbh!]
\centering
\caption{
Description of four real datasets.
For each dataset, the sample size ($n$), the proportion of class 0 observations ($n_0/n$), the number of features ($d$), and the class encodings are listed.
\label{tab:s6-data}
}
\begin{tabular}{rrrrrr}
\hline
 dataset & $n$ & $n_0/n$ & $d$ & class $0$ & class $1$ \\
 \hline
 \texttt{Diabetes} & $768$ & $0.35$ & $8$ & diabetes & negative \\
 \texttt{Thyroid} & $3090$ & $0.09$ & $18$ & euthyroid sick & negative \\
 \texttt{Breast cancer} & $960$ & $0.46$ & $5$ & malignant & benign \\
 \texttt{Marketing} & $4119$ & $0.11$ & $18$ & yes & no \\
\hline
\end{tabular}
\end{table}

\begin{figure}[tbh!]
\centering
\includegraphics[width=.8\textwidth]{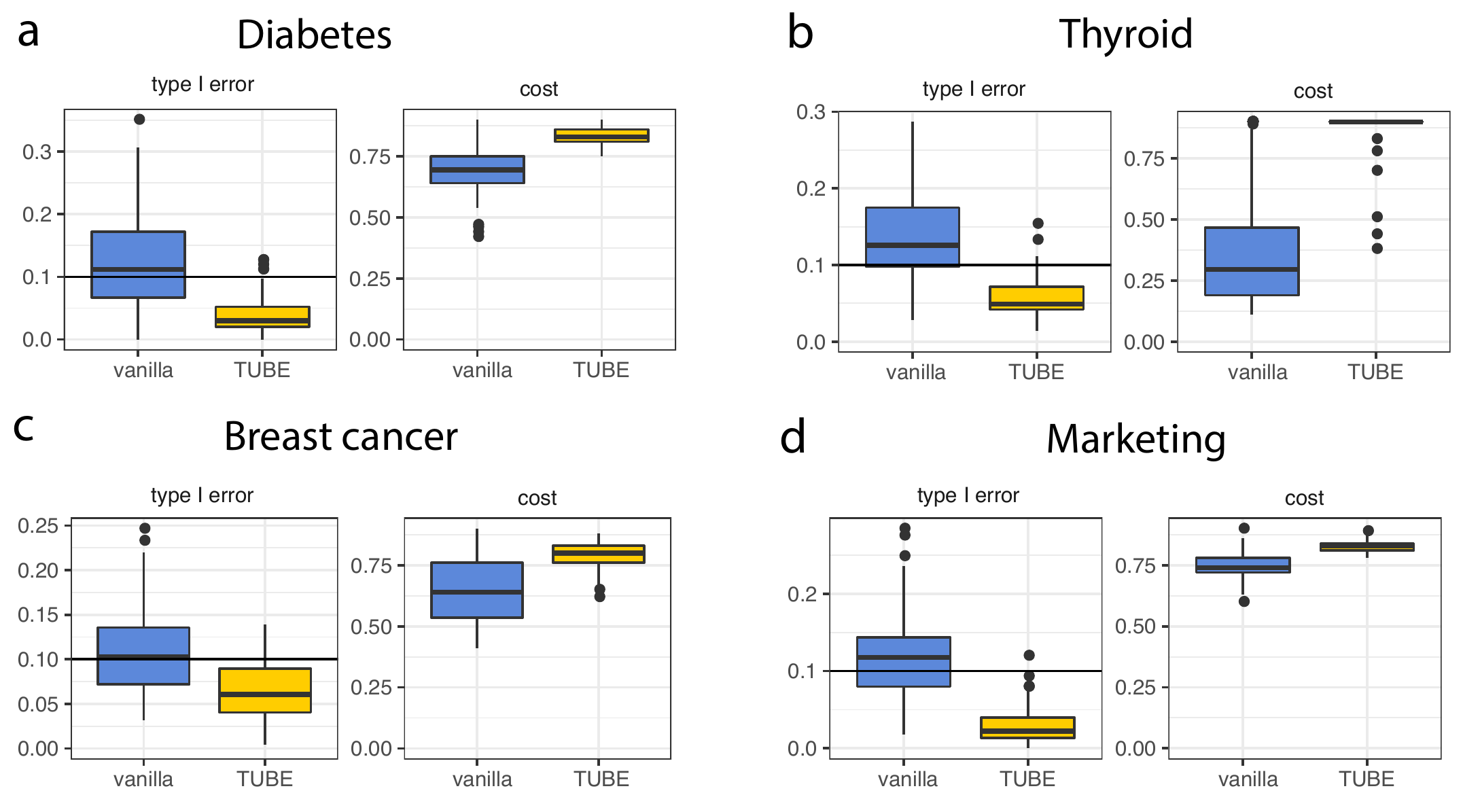}
\caption{
Comparison of the vanilla-CS (``vanilla'') and TUBE-CS (``TUBE'') algorithms on the four real datasets (Table \ref{tab:s6-data}): (a) \texttt{Diabetes}, (b) \texttt{Throid}, (c) \texttt{Breast cancer}, (d) and \texttt{Marketing}. The type I errors are calculated on evaluation data across $50$ random splits, each of which divides each dataset into two halves for training and evaluation. Boxplots are shown for the type I errors and the type I error costs ($c_0$) of the classifiers constructed by each algorithm.
\label{fig:s6-LG}
}
\end{figure}

For each real dataset, we randomly split the data into two subsets of equal size, one used as the training sample for constructing CS classifiers by the vanilla-CS or the TUBE-CS algorithm, and the other used as the evaluation sample for calculating the type I and II errors. When implementing the vanilla-CS and TUBE-CS algorithms, we use the logistic regression as the classification method and the stratification approach as the CS implementation. We set the target type I error upper bound to $\alpha = 0.05$ and the violation rate to $\delta = 10\%$, and we repeat the random splitting for $50$ times. Since we do not have a large evaluation set to approximate the population in these real datasets, we cannot assess the violation rates of CS classifiers' population type I errors. However, using the empirical type I errors on the evaluation sample as a proxy, we can see that the TUBE-CS classifiers clearly have better control of type I error than the vanilla-CS classifiers do on all the four datasets (Figure \ref{fig:s6-LG}). 

These results suggest that the TUBE-CS algorithm is a useful tool for choosing the type I error cost $c_0$ in CS classification, when users have a target type I error upper bound in mind. Moreover, the TUBE algorithm offers a way to estimate an upper bound on the type I error of a CS classifier constructed with a cost $c_0$, thus improving the interpretability of CS classifiers.


\section{Conclusions}
In this article, we discuss two classification paradigms, CS and NNP, for prioritizing misclassification errors in asymmetric binary classification problems. Our work, for the first time, discusses the methodological connections between the two paradigms. We identify two special cases in which we can construct a CS classifier that is identical to a given NP classifier. For a given CS classifier, using a left-out class 0 sample, the TUBEc algorithm delivers a numerically validated estimate of a high-probability upper bound on the population type I error of the CS classifier.  Moreover, the TUBE algorithm, which uses only the training data and a pre-specified type I error cost (without having access to an actual CS classifier), delivers a less accurate but still reasonable estimate of the upper bound. The TUBE algorithm is valuable when the overall class 0 (usually the more severe state) sample size is small. Powered by the TUBE algorithm, the TUBE-CS algorithm offers a substitute for the NP umbrella algorithm for constructing classifiers so that the classifiers have population type I errors under a pre-specified upper bound $\alpha$ with high probability. The TUBE-CS algorithm is preferable when the high-probability requirement is not strict or the minimum class $0$ sample size required by the NP umbrella algorithm is not met.

\section{Code availability}
The source code for the statistical analysis in this work is available at \url{https://github.com/Vivianstats/TUBE}.

\newpage

\appendix
\section*{Appendix 1. Supplementary Text}
\label{app:theorem}

\setcounter{table}{0}
\renewcommand{\thetable}{A\arabic{table}}
\setcounter{figure}{0}
\renewcommand{\thefigure}{A\arabic{figure}}
\setcounter{algocf}{0}
\renewcommand{\thealgocf}{A\arabic{algocf}}

\subsection*{Proof of Proposition 2}
\setcounter{prop}{1}

\begin{proof}
When the type I error cost $c_0$ takes the form\begin{align*}
c_0 
&= 
\frac{t_\text{NP}\hat\pi_0}{(1-\hat\pi_0)(1-t_\text{NP}) + t_\text{NP}\hat\pi_0}\;,
\end{align*}
in view of our assumption $c_0 + c_1 = 1$,  we have $\forall X \in \mathcal{X}$,
\begin{align*}
&\tilde \eta(X) > \frac{1}{2}\\
&\Leftrightarrow
\frac{\hat f_1(X)}{\hat f_0(X)} > \frac{c_0}{1-c_0}\\
&\Leftrightarrow
\frac{\hat f_1(X)}{\hat f_0(X)} > 
\frac{\frac{t_\text{NP}\hat\pi_0}{(1-\hat\pi_0)(1-t_\text{NP}) + t_\text{NP}\hat\pi_0}}
{\frac{(1-\hat\pi_0)(1-t_\text{NP})}{(1-\hat\pi_0)(1-t_\text{NP}) + t_\text{NP}\hat\pi_0}}\\
&\Leftrightarrow
\frac{\hat f_1(X)}{\hat f_0(X)} > \frac{t_\text{NP}\hat\pi_0}{(1-\hat\pi_0)(1-t_\text{NP})}\\
&\Leftrightarrow
\hat \eta(X) > t_\text{NP}\;.
\end{align*}
Therefore, the rebalancing CS classifier $\hat \phi_\text{CS}^{\text{r}}(X) = \1(\tilde \eta(X)> 1/2)$ is the same as the NP classifier $\hat\phi_\text{NP}(X) = \1(\hat \eta(X)> t_\text{NP})$.
\end{proof}

\subsection*{Proof of Proposition 4}
\setcounter{prop}{3}

\begin{proof}
We consider the type I errors of the CS classifier $\hat \phi_\text{CS}(\cdot) = \1(s(\cdot)>t_\text{CS})$ and the surrogate classifier $\hat \phi_{k^*_s}(\cdot) = \1(s(\cdot)>T_{(k^*_s)})$. Given that $\hat \phi_\text{CS}$ is fixed, its population type I error is a fixed value 
$$R_0(\hat \phi_\text{CS}) = 1-F(t_\text{CS})\,,$$
while the population type I error of $\hat \phi_{k^*_s}$ is a random variable 
$$R_0(\hat \phi_{k^*_s}) = 1-F(T_{(k^*_s)})$$
due to the randomness of $T_{(k^*_s)}$.

Under the assumption that $T_{(1)}\leq t_{\text{CS}}$, 
we have $T_{(k^*_s)} \leq t_\text{CS}$, and hence $R_0(\hat \phi_{k^*_s}) \geq R_0(\hat \phi_\text{CS})$. In other words, the population type I error of the CS classifier is no greater than the population type I error of the surrogate classifier.

Next, we calculate the probability (regarding measure $\bbP_m$) that the population type I error of the surrogate classifier $\hat \phi_{k^*_s}$ exceeds $\alpha$:
\begin{align*}
\bbP_m(R_0(\hat \phi_{k^*_s}) > \alpha) 
&= \bbP_m(1-F(T_{(k^*_s)})> \alpha) \\
&= \bbP_m(T_{(k^*_s)}<F^{-1}(1-\alpha)) \\
&= \sum_{k=1}^{m}\bbP_m(T_{(k)}<F^{-1}(1-\alpha),\; k^*_s = k) \\
&= \sum_{k=1}^{m-1}\bbP_m(T_{(k)}<F^{-1}(1-\alpha),\; T_{(k)}\leq t_\text{CS},\; T_{(k+1)}>t_\text{CS}) \\
&\;\; + \bbP_m(T_{(m)}<F^{-1}(1-\alpha),\; T_{(m)}\leq t_\text{CS})\;.
\end{align*}
If $t_\text{CS} < F^{-1}(1-\alpha)$,
\begin{align*}
\bbP_m(R_0(\hat \phi_{k^*_s}) > \alpha)
&=\sum_{k=1}^{m-1}\bbP_m(T_{(k)}\leq t_\text{CS},\; T_{(k+1)}>t_\text{CS}) + \bbP_m(T_{(m)}\leq t_\text{CS}) = 1\;.
\end{align*}
If $t_\text{CS} \geq F^{-1}(1-\alpha)$,
\allowdisplaybreaks
\begin{align*}
\bbP_m(R_0(\hat \phi_{k^*_s}) > \alpha)
&= \sum_{k=1}^{m-1}\bbP_m(T_{(k)}<F^{-1}(1-\alpha),\; T_{(k+1)}>t_\text{CS}) + \bbP_m(T_{(m)}<F^{-1}(1-\alpha))\\
&= \sum_{k=1}^{m-1}{m\choose k}\bbP_m(T_1<F^{-1}(1-\alpha),\; \dots,\; T_k<F^{-1}(1-\alpha),\; T_{k+1}>t_\text{CS},\;\dots,\;T_{m-1}>t_\text{CS}) \\
&  + \bbP_m(T_1<F^{-1}(1-\alpha),\; \dots,\; T_m<F^{-1}(1-\alpha))\\
&\leq \sum_{k=1}^{m}{m\choose k}(1-\alpha)^k\left(1-F(t_\text{CS})\right)^{m-k}\\
&\overset{u = 1-F(t_\text{CS})}{=}(1-\alpha+u)^m - u^m\;.
\end{align*}
In summary,
\begin{align*}
\bbP_m(R_0(\hat \phi_{k^*_s})>\alpha)
\begin{cases}
=1 & \text{if}\ t_\text{CS} < F^{-1}(1-\alpha)\\
\leq (1-\alpha+u)^m- u^m & \text{if}\ t_\text{CS} \geq F^{-1}(1-\alpha)
\end{cases}
\;.
\end{align*}
In other words, when $t_\text{CS} < F^{-1}(1-\alpha)$, the type I error of $\hat \phi_{k^*_s}$ exceeds $\alpha$ with probability $1$; 
when $t_\text{CS} \geq F^{-1}(1-\alpha)$, the type I error of $\hat \phi_{k^*_s}$ exceeds $\alpha$ with probability no larger than $\delta_s = (2-\alpha-F(t_\text{CS}))^m-(1-F(t_\text{CS}))^m$. 
\end{proof}

\subsection*{Algorithm A1}

\begin{algorithm}[htb!]
\DontPrintSemicolon
\caption{\label{alm:TUBEc-cs}
TUBEc-assisted CS classification (TUBEc-CS)}
\SetKw{KwBy}{by}
\SetKwInOut{Input}{Input}\SetKwInOut{Output}{Output}
\SetAlgoLined

\Input{$\mathcal{S}=\mathcal{S}^0\cup \mathcal{S}^1$: training data ($\mathcal{S}^0$ for class $0$ and $\mathcal{S}^1$ for class $1$)\\
$\alpha$: target upper bound on the type I error\\
$c_{0,1} < c_{0,2} < \dots < c_{0,I}$: candidate type I error costs\\
$\delta$: violation rate
}
$\mathcal{S}^0_1, \mathcal{S}^0_2\leftarrow$ randomly split class 0 data $\mathcal{S}^0$\;

\For{$i\gets1$ \KwTo $I$}{
  $\hat\phi_i(\cdot) = $ CS-classifier($\mathcal{S}^1\cup\mathcal{S}^0_1, c_{0, i}$)\\
  $\hat\alpha_i^{\textrm{c}}$ = TUBEc($\hat \phi_i(\cdot), \mathcal{S}^0_2, \delta$)
}
$i^{*\textrm{c}} = \min\{i: \hat\alpha_i^{\textrm{c}}\leq \alpha\}$\;

\Output{$\hat{\phi}^{\textrm{TUBEc-CS}}(\cdot) =\hat\phi_{i^{*\textrm{c}}}(\cdot)$\tcp*{the TUBE-CS classifier}}
\end{algorithm}

%
%
%

\clearpage

\section*{Appendix 2: Supplementary Figures}




\begin{figure}[tbh!]
\centering
\includegraphics[width=.9\textwidth]{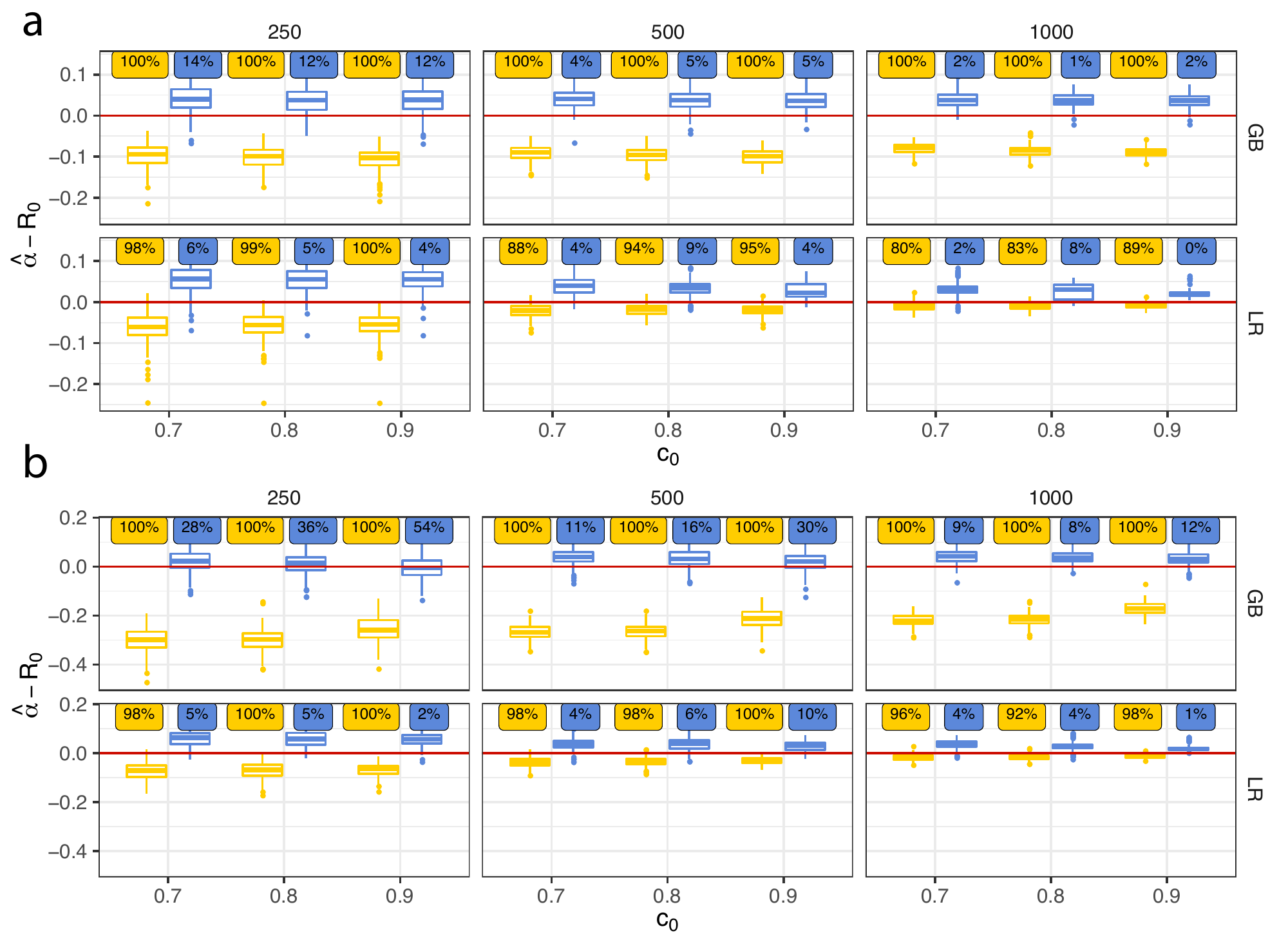}
\caption{Performance of the TUBE estimator ($\hat\alpha^\text{TUBE}$; blue) and the empirical estimator ($\hat\alpha^\text{emp}$; yellow).
Data are generated from (a) the Multivariate $t$ distribution and (b) the Mixture distribution (Section~\ref{sec: cs-sim}). We use the stratification approach with two classification methods, gradient boosting (GB) and logistic regression (LR), to construct CS classifiers.
The boxplots show the distributions of the difference between the estimated type I error upper bound ($\hat\alpha$) and the actual population type I error of the given CS classifier ($R_0$) (approximated on a large sample of size $10^6$).
The horizontal axis denotes the type I error cost ($c_0 = 0.7$, $0.8$, or $0.9$). The training sample size ($250$, $500$, or $1000$) is labeled on the top of each column.
The violation rate (percentage of simulations in which $R_0 > \hat\alpha$) is labeled on the top of each boxplot.
\label{fig:tube-supp}
}
\end{figure}

\clearpage
\bibliography{npcs}

\begin{thebibliography}{53}
\providecommand{\natexlab}[1]{#1}
\providecommand{\url}[1]{\texttt{#1}}
\expandafter\ifx\csname urlstyle\endcsname\relax
  \providecommand{\doi}[1]{doi: #1}\else
  \providecommand{\doi}{doi: \begingroup \urlstyle{rm}\Url}\fi

\bibitem[Ali et~al.(2016)Ali, Majid, Javed, and Sattar]{ali2016can}
Safdar Ali, Abdul Majid, Syed~Gibran Javed, and Mohsin Sattar.
\newblock Can-csc-gbe: Developing cost-sensitive classifier with gentleboost
  ensemble for breast cancer classification using protein amino acids and
  imbalanced data.
\newblock \emph{Computers in Biology and Medicine}, 73:\penalty0 38--46, 2016.

\bibitem[Alizadehsani et~al.(2012)Alizadehsani, Hosseini, Sani, Ghandeharioun,
  and Boghrati]{alizadehsani2012diagnosis}
Roohallah Alizadehsani, Mohammad~Javad Hosseini, Zahra~Alizadeh Sani, Asma
  Ghandeharioun, and Reihane Boghrati.
\newblock Diagnosis of coronary artery disease using cost-sensitive algorithms.
\newblock In \emph{Data Mining Workshops (ICDMW), 2012 IEEE 12th International
  Conference on}, pages 9--16. IEEE, 2012.

\bibitem[Artan et~al.(2010)Artan, Haider, Langer, Van~der Kwast, Evans, Yang,
  Wernick, Trachtenberg, and Yetik]{artan2010prostate}
Yusuf Artan, Masoom~A Haider, Deanna~L Langer, Theodorus~H Van~der Kwast,
  Andrew~J Evans, Yongyi Yang, Miles~N Wernick, John Trachtenberg, and
  Imam~Samil Yetik.
\newblock Prostate cancer localization with multispectral mri using
  cost-sensitive support vector machines and conditional random fields.
\newblock \emph{IEEE Transactions on Image Processing}, 19\penalty0
  (9):\penalty0 2444--2455, 2010.

\bibitem[Bahnsen et~al.(2013)Bahnsen, Stojanovic, Aouada, and
  Ottersten]{bahnsen2013cost}
Alejandro~Correa Bahnsen, Aleksandar Stojanovic, Djamila Aouada, and Bj{\"o}rn
  Ottersten.
\newblock Cost sensitive credit card fraud detection using bayes minimum risk.
\newblock In \emph{2013 12th International Conference on Machine Learning and
  Applications}, volume~1, pages 333--338. IEEE, 2013.

\bibitem[Beck et~al.(2000)Beck, King, and Zeng]{beck2000improving}
Nathaniel Beck, Gary King, and Langche Zeng.
\newblock Improving quantitative studies of international conflict: A
  conjecture.
\newblock \emph{American Political Science Review}, 94\penalty0 (1):\penalty0
  21--35, 2000.

\bibitem[Bradford et~al.(1998)Bradford, Kunz, Kohavi, Brunk, and
  Brodley]{bradford1998pruning}
Jeffrey~P Bradford, Clayton Kunz, Ron Kohavi, Cliff Brunk, and Carla~E Brodley.
\newblock Pruning decision trees with misclassification costs.
\newblock In \emph{European Conference on Machine Learning}, pages 131--136.
  Springer, 1998.

\bibitem[Breiman et~al.(1984)Breiman, Friedman, Stone, and
  Olshen]{breiman2017classification}
Leo Breiman, Jerome Friedman, Charles~J. Stone, and R.A. Olshen.
\newblock \emph{Classification and regression trees}.
\newblock Routledge, 1984.

\bibitem[Carreras and Marquez(2001)]{carreras2001boosting}
Xavier Carreras and Lluis Marquez.
\newblock Boosting trees for anti-spam email filtering.
\newblock \emph{arXiv preprint cs/0109015}, 2001.

\bibitem[Cederman and Weidmann(2017)]{cederman2017predicting}
Lars-Erik Cederman and Nils~B Weidmann.
\newblock Predicting armed conflict: Time to adjust our expectations?
\newblock \emph{Science}, 355\penalty0 (6324):\penalty0 474--476, 2017.

\bibitem[Chan and Stolfo(1998)]{chan1998toward}
Philip~K Chan and Salvatore~J Stolfo.
\newblock Toward scalable learning with non-uniform class and cost
  distributions: A case study in credit card fraud detection.
\newblock In \emph{International Conference on Knowledge Discovery and Data
  Mining}, volume~98, pages 164--168, 1998.

\bibitem[Chen et~al.(2004)Chen, Liaw, and Breiman]{chen2004using}
C~Chen, A~Liaw, and L~Breiman.
\newblock Using random forest to learn imbalanced data. statistics department
  of university of california at berkeley.
\newblock Technical report, Berkeley. Technical Report 666, 2004.

\bibitem[Domingos(1999)]{domingos1999metacost}
Pedro Domingos.
\newblock Metacost: A general method for making classifiers cost-sensitive.
\newblock In \emph{Proceedings of the fifth ACM SIGKDD International Conference
  on Knowledge Discovery and Data Mining}, pages 155--164. ACM, 1999.

\bibitem[Drummond and Holte(2000)]{drummond2000exploiting}
Chris Drummond and Robert~C. Holte.
\newblock Exploiting the cost (in)sensitivity of decision tree splitting
  criteria.
\newblock In \emph{Proceedings of the Seventeenth International Conference on
  Machine Learning}, ICML '00, page 239–246, San Francisco, CA, USA, 2000.
  Morgan Kaufmann Publishers Inc.
\newblock ISBN 1558607072.

\bibitem[Dua and Graff(2017)]{Dua:2019}
Dheeru Dua and Casey Graff.
\newblock {UCI} machine learning repository, 2017.
\newblock URL \url{http://archive.ics.uci.edu/ml}.

\bibitem[Duda et~al.(1973)Duda, Hart, Stork, et~al.]{duda1973pattern}
Richard~O Duda, Peter~E Hart, David~G Stork, et~al.
\newblock \emph{Pattern classification}, volume~2.
\newblock Wiley New York, 1973.

\bibitem[Elder~IV(1996)]{elder1996machine}
John~F Elder~IV.
\newblock Machine learning, neural, and statistical classification, 1996.

\bibitem[Fan et~al.(1999)Fan, Stolfo, Zhang, and Chan]{fan1999adacost}
Wei Fan, Salvatore~J Stolfo, Junxin Zhang, and Philip~K Chan.
\newblock Adacost: misclassification cost-sensitive boosting.
\newblock In \emph{International Conference on Machine Learning}, volume~99,
  pages 97--105, 1999.

\bibitem[Fern{\'a}ndez-G{\'o}mez et~al.(2017)Fern{\'a}ndez-G{\'o}mez,
  Asencio-Cort{\'e}s, Troncoso, and
  Mart{\'\i}nez-{\'A}lvarez]{fernandez2017large}
Manuel~Jes{\'u}s Fern{\'a}ndez-G{\'o}mez, Gualberto Asencio-Cort{\'e}s, Alicia
  Troncoso, and Francisco Mart{\'\i}nez-{\'A}lvarez.
\newblock Large earthquake magnitude prediction in chile with imbalanced
  classifiers and ensemble learning.
\newblock \emph{Applied Sciences}, 7\penalty0 (6):\penalty0 625, 2017.

\bibitem[Friedman et~al.(2010)Friedman, Hastie, and Tibshirani]{glmnet}
Jerome Friedman, Trevor Hastie, and Robert Tibshirani.
\newblock Regularization paths for generalized linear models via coordinate
  descent.
\newblock \emph{Journal of Statistical Software}, 33\penalty0 (1):\penalty0
  1--22, 2010.
\newblock URL \url{http://www.jstatsoft.org/v33/i01/}.

\bibitem[Horrocks et~al.(2015)Horrocks, Wedge, Holden, Kovesi, Clarke, and
  Vann]{horrocks2015classification}
Tom Horrocks, Daniel Wedge, Eun-Jung Holden, Peter Kovesi, Nick Clarke, and
  John Vann.
\newblock Classification of gold-bearing particles using visual cues and
  cost-sensitive machine learning.
\newblock \emph{Mathematical Geosciences}, 47\penalty0 (5):\penalty0 521--545,
  2015.

\bibitem[King et~al.(1995)King, Feng, and Sutherland]{king1995statlog}
Ross~D. King, Cao Feng, and Alistair Sutherland.
\newblock Statlog: comparison of classification algorithms on large real-world
  problems.
\newblock \emph{Applied Artificial Intelligence an International Journal},
  9\penalty0 (3):\penalty0 289--333, 1995.

\bibitem[Krawczyk et~al.(2015)Krawczyk, Schaefer, and
  Wo{\'z}niak]{krawczyk2015hybrid}
Bartosz Krawczyk, Gerald Schaefer, and Micha{\l} Wo{\'z}niak.
\newblock A hybrid cost-sensitive ensemble for imbalanced breast thermogram
  classification.
\newblock \emph{Artificial Intelligence in Medicine}, 65\penalty0 (3):\penalty0
  219--227, 2015.

\bibitem[Kukar et~al.(1998)Kukar, Kononenko, et~al.]{kukar1998cost}
Matjaz Kukar, Igor Kononenko, et~al.
\newblock Cost-sensitive learning with neural networks.
\newblock In \emph{European Conference on Artificial Intelligence}, pages
  445--449, 1998.

\bibitem[Lan et~al.(2010)Lan, Hu, Patuwo, and Zhang]{lan2010investigation}
Jyhshyan Lan, Michael~Y Hu, Eddy Patuwo, and G~Peter Zhang.
\newblock An investigation of neural network classifiers with unequal
  misclassification costs and group sizes.
\newblock \emph{Decision Support Systems}, 48\penalty0 (4):\penalty0 582--591,
  2010.

\bibitem[Liu et~al.(2016)Liu, Lu, Yan, Xia, and An]{liu2016applying}
Yanqiu Liu, Huijuan Lu, Ke~Yan, Haixia Xia, and Chunlin An.
\newblock Applying cost-sensitive extreme learning machine and dissimilarity
  integration to gene expression data classification.
\newblock \emph{Computational Intelligence and Neuroscience}, 2016, 2016.

\bibitem[Lu and Wang(2008)]{lu2008ground}
Wei-Zhen Lu and Dong Wang.
\newblock Ground-level ozone prediction by support vector machine approach with
  a cost-sensitive classification scheme.
\newblock \emph{Science of the Total Environment}, 395\penalty0 (2-3):\penalty0
  109--116, 2008.

\bibitem[Majka(2018)]{naivebayes}
Michal Majka.
\newblock \emph{naivebayes: High Performance Implementation of the Naive Bayes
  Algorithm}, 2018.
\newblock URL \url{https://CRAN.R-project.org/package=naivebayes}.
\newblock R package version 0.9.2.

\bibitem[Margineantu(2000)]{margineantu2000does}
Dragos Margineantu.
\newblock When does imbalanced data require more than cost-sensitive learning.
\newblock In \emph{Proceedings of the AAAI 2000 Workshop on Learning from
  Imbalanced Data Sets}, pages 47--50, 2000.

\bibitem[Margineantu(2002)]{margineantu2002class}
Dragos~D Margineantu.
\newblock Class probability estimation and cost-sensitive classification
  decisions.
\newblock In \emph{European Conference on Machine Learning}, pages 270--281.
  Springer, 2002.

\bibitem[Mazurowski et~al.(2008)Mazurowski, Habas, Zurada, Lo, Baker, and
  Tourassi]{mazurowski2008training}
Maciej~A Mazurowski, Piotr~A Habas, Jacek~M Zurada, Joseph~Y Lo, Jay~A Baker,
  and Georgia~D Tourassi.
\newblock Training neural network classifiers for medical decision making: The
  effects of imbalanced datasets on classification performance.
\newblock \emph{Neural Networks}, 21\penalty0 (2-3):\penalty0 427--436, 2008.

\bibitem[Meyer et~al.(2017)Meyer, Dimitriadou, Hornik, Weingessel, and
  Leisch]{e1071}
David Meyer, Evgenia Dimitriadou, Kurt Hornik, Andreas Weingessel, and
  Friedrich Leisch.
\newblock \emph{e1071: Misc Functions of the Department of Statistics,
  Probability Theory Group (Formerly: E1071), TU Wien}, 2017.
\newblock URL \url{https://CRAN.R-project.org/package=e1071}.
\newblock R package version 1.6-8.

\bibitem[Moon et~al.(2012)Moon, Shen, Bae, Huang, Chen, and
  Chang]{moon2012computer}
Woo~Kyung Moon, Yi-Wei Shen, Min~Sun Bae, Chiun-Sheng Huang, Jeon-Hor Chen, and
  Ruey-Feng Chang.
\newblock Computer-aided tumor detection based on multi-scale blob detection
  algorithm in automated breast ultrasound images.
\newblock \emph{IEEE Transactions on Medical Imaging}, 32\penalty0
  (7):\penalty0 1191--1200, 2012.

\bibitem[Park et~al.(2011)Park, Chun, and Kim]{park2011cost}
Yoon-Joo Park, Se-Hak Chun, and Byung-Chun Kim.
\newblock Cost-sensitive case-based reasoning using a genetic algorithm:
  Application to medical diagnosis.
\newblock \emph{Artificial Intelligence in Medicine}, 51\penalty0 (2):\penalty0
  133--145, 2011.

\bibitem[Pelayo and Dick(2007)]{pelayo2007applying}
Lourdes Pelayo and Scott Dick.
\newblock Applying novel resampling strategies to software defect prediction.
\newblock In \emph{Fuzzy Information Processing Society, 2007. NAFIPS'07.
  Annual Meeting of the North American}, pages 69--72. IEEE, 2007.

\bibitem[Pelayo and Dick(2012)]{pelayo2012evaluating}
Lourdes Pelayo and Scott Dick.
\newblock Evaluating stratification alternatives to improve software defect
  prediction.
\newblock \emph{IEEE Transactions on Reliability}, 61\penalty0 (2):\penalty0
  516--525, 2012.

\bibitem[{R Core Team}(2013)]{stats}
{R Core Team}.
\newblock \emph{R: A Language and Environment for Statistical Computing}.
\newblock R Foundation for Statistical Computing, Vienna, Austria, 2013.
\newblock URL \url{http://www.R-project.org/}.
\newblock {ISBN} 3-900051-07-0.

\bibitem[Rigollet and Tong(2011)]{rigollet2011neyman}
Philippe Rigollet and Xin Tong.
\newblock Neyman-pearson classification, convexity and stochastic constraints.
\newblock \emph{The Journal of Machine Learning Research}, 12:\penalty0
  2831--2855, 2011.

\bibitem[Sahin et~al.(2013)Sahin, Bulkan, and Duman]{sahin2013cost}
Yusuf Sahin, Serol Bulkan, and Ekrem Duman.
\newblock A cost-sensitive decision tree approach for fraud detection.
\newblock \emph{Expert Systems with Applications}, 40\penalty0 (15):\penalty0
  5916--5923, 2013.

\bibitem[Schaefer et~al.(2007)Schaefer, Nakashima, Yokota, and
  Ishibuchi]{schaefer2007cost}
Gerald Schaefer, Tomoharu Nakashima, Yasuyuki Yokota, and Hisao Ishibuchi.
\newblock Cost-sensitive fuzzy classification for medical diagnosis.
\newblock In \emph{Computational Intelligence and Bioinformatics and
  Computational Biology, 2007. CIBCB'07. IEEE Symposium on}, pages 312--316.
  IEEE, 2007.

\bibitem[Scott(2005)]{scott2005comparison}
C~Scott.
\newblock Comparison and design of neyman-pearson classifiers, 2005.
\newblock URL \url{http://www.stat.rice.edu/~cscott/pubs/npdesign.pdf}.

\bibitem[Scott and Nowak(2005)]{scott2005neyman}
Clayton Scott and Robert Nowak.
\newblock A neyman-pearson approach to statistical learning.
\newblock \emph{IEEE Transactions on Information Theory}, 51\penalty0
  (11):\penalty0 3806--3819, 2005.

\bibitem[Sun et~al.(2007)Sun, Kamel, Wong, and Wang]{sun2007cost}
Yanmin Sun, Mohamed~S Kamel, Andrew~KC Wong, and Yang Wang.
\newblock Cost-sensitive boosting for classification of imbalanced data.
\newblock \emph{Pattern Recognition}, 40\penalty0 (12):\penalty0 3358--3378,
  2007.

\bibitem[Tong et~al.(2018)Tong, Feng, and Li]{tong2018neyman}
Xin Tong, Yang Feng, and Jingyi~Jessica Li.
\newblock Neyman-pearson classification algorithms and np receiver operating
  characteristics.
\newblock \emph{Science Advances}, 4\penalty0 (2):\penalty0 eaao1659, 2018.

\bibitem[Tong et~al.(2020)Tong, Xia, Wang, and Feng]{Tong.Xia.Wang.Feng.2020}
Xin Tong, Lucy Xia, Jiacheng Wang, and Yang Feng.
\newblock Neyman-pearson classification: parametrics and sample size
  requirement.
\newblock \emph{Journal of Machine Learning Research}, 21\penalty0
  (12):\penalty0 1--48, 2020.

\bibitem[Turney(2002)]{turney2002types}
Peter~D Turney.
\newblock Types of cost in inductive concept learning.
\newblock \emph{arXiv preprint cs/0212034}, 2002.

\bibitem[Venables and Ripley(2002)]{mass}
W.~N. Venables and B.~D. Ripley.
\newblock \emph{Modern Applied Statistics with S}.
\newblock Springer, New York, fourth edition, 2002.
\newblock URL \url{http://www.stats.ox.ac.uk/pub/MASS4}.
\newblock ISBN 0-387-95457-0.

\bibitem[Vidrighin and Potolea(2008)]{vidrighin2008proicet}
Camelia Vidrighin and Rodica Potolea.
\newblock Proicet: a cost-sensitive system for prostate cancer data.
\newblock \emph{Health Informatics Journal}, 14\penalty0 (4):\penalty0
  297--307, 2008.

\bibitem[Webb and Ting(2005)]{webb2005application}
Geoffrey~I Webb and Kai~Ming Ting.
\newblock On the application of roc analysis to predict classification
  performance under varying class distributions.
\newblock \emph{Machine Learning}, 58\penalty0 (1):\penalty0 25--32, 2005.

\bibitem[Xia et~al.(2020)Xia, Zhao, Wu, and Tong]{xia2020intentional}
Lucy Xia, Richard Zhao, Yanhui Wu, and Xin Tong.
\newblock Intentional control of type i error over unconscious data distortion:
  A neyman--pearson approach to text classification.
\newblock \emph{Journal of the American Statistical Association}, pages 1--14,
  2020.

\bibitem[Zadrozny et~al.(2003)Zadrozny, Langford, and Abe]{zadrozny2003cost}
Bianca Zadrozny, John Langford, and Naoki Abe.
\newblock Cost-sensitive learning by cost-proportionate example weighting.
\newblock In \emph{Data Mining, 2003. ICDM 2003. Third IEEE International
  Conference on}, pages 435--442. IEEE, 2003.

\bibitem[Zhao et~al.(2016)Zhao, Feng, Wang, and Tong]{zhao2016neyman}
Anqi Zhao, Yang Feng, Lie Wang, and Xin Tong.
\newblock Neyman-pearson classification under high-dimensional settings.
\newblock \emph{The Journal of Machine Learning Research}, 17\penalty0
  (1):\penalty0 7469--7507, 2016.

\bibitem[Zhou et~al.(2014)Zhou, Yao, and Luo]{zhou2014cost}
Bing Zhou, Yiyu Yao, and Jigang Luo.
\newblock Cost-sensitive three-way email spam filtering.
\newblock \emph{Journal of Intelligent Information Systems}, 42\penalty0
  (1):\penalty0 19--45, 2014.

\bibitem[Zhou and Liu(2006)]{zhou2006training}
Zhi-Hua Zhou and Xu-Ying Liu.
\newblock Training cost-sensitive neural networks with methods addressing the
  class imbalance problem.
\newblock \emph{IEEE Transactions on Knowledge and Data Engineering},
  18\penalty0 (1):\penalty0 63--77, 2006.

\end{thebibliography}

\end{document}